\pdfoutput=1

\documentclass[11pt]{article}

\usepackage[]{ACL2023}

\usepackage{times}
\usepackage{latexsym}

\usepackage[T1]{fontenc}

\usepackage[utf8]{inputenc}

\usepackage{microtype}

\usepackage{inconsolata}

\usepackage{tabularx} 
\usepackage{array}
\usepackage{multirow}
\usepackage{booktabs}
\usepackage{graphicx} 
\usepackage{url} 
\usepackage{amsthm,amsmath,amssymb} 
\usepackage{mathrsfs} 
\usepackage{bm} 
\usepackage{pifont} 
\usepackage{float}
\usepackage{lipsum}
\usepackage{soul}
\usepackage{caption}
\usepackage{subcaption}
\usepackage{enumitem}
\usepackage{bbm}
\usepackage{dsfont}
\usepackage{color}
\usepackage{colortbl}
\usepackage{xcolor}
\definecolor{deepred}{RGB}{180,0,0} 
\usepackage{xspace}

\title{Learning to Edit: Aligning LLMs with Knowledge Editing}

\author{
Yuxin Jiang$^{1,2}$\thanks{~~Work done during the internship at Huawei Noah's Ark Lab. Data and code: \url{https://github.com/YJiangcm/LTE}.}, 
Yufei Wang$^{3}$, 
Chuhan Wu$^{3}$, 
Wanjun Zhong$^{3}$, 
\\
\textbf{Xingshan Zeng}$^{3}$\textbf{,} 
\textbf{Jiahui Gao}$^{3}$\textbf{,} 
\textbf{Liangyou Li}$^{3}$\textbf{,} 
\textbf{Xin Jiang}$^{3}$\textbf{,} 
\\
\textbf{Lifeng Shang}$^{3}$\textbf{,} 
\textbf{Ruiming Tang}$^{3}$\textbf{,} 
\textbf{Qun Liu}$^{3}$\textbf{,} 
\textbf{Wei Wang}$^{1,2}$
\\
The Hong Kong University of Science and Technology (Guangzhou)$^1$, \\
The Hong Kong University of Science and Technology$^2$, Huawei Noah’s Ark Lab$^3$ \\
yjiangcm@connect.ust.hk, wang.yufei1@huawei.com, weiwcs@ust.hk 
}

\begin{document}
\maketitle

\begin{abstract}
Knowledge editing techniques, aiming to efficiently modify a minor proportion of knowledge in large language models (LLMs) without negatively impacting performance across other inputs, have garnered widespread attention.
However, existing methods predominantly rely on memorizing the updated knowledge, impeding LLMs from effectively combining the new knowledge with their inherent knowledge when answering questions.
To this end, we propose a \textit{Learning to Edit} (LTE) framework, focusing on teaching LLMs to \textbf{apply} updated knowledge into input questions,
inspired by the philosophy of ``\textit{Teach a man to fish}.''
LTE features a two-phase process: (i) the Alignment Phase, which fine-tunes LLMs on a meticulously curated parallel dataset to make reliable, in-scope edits while preserving out-of-scope information and linguistic proficiency;
and (ii) the Inference Phase, which employs a retrieval-based mechanism for real-time and mass knowledge editing.
By comparing our approach with seven advanced baselines across four popular knowledge editing benchmarks and two LLM architectures, we demonstrate LTE's superiority in knowledge editing performance, robustness in both batch and sequential editing, minimal interference on general tasks, and rapid editing speeds.
\end{abstract}

\section{Introduction}
The transformative potential of large language models (LLMs)~\cite{brown2020fewshot, 2023gpt4, touvron2023llama} has been unequivocally underscored by their unparalleled efficacy across a myriad of applications~\cite{chen2021code, openai2022chatgpt, 2023gpt4}.
Nonetheless, the dynamic nature of the world necessitates frequent updates to LLMs to rectify outdated information or integrate new knowledge, thereby safeguarding their sustained pertinence.
Naively training a new LLM from scratch to incorporate updated knowledge could result in substantial computational overhead and is frequently deemed impractical.
To this end, the concept of \textbf{knowledge editing} has been introduced~\cite{DBLP:conf/iclr/SinitsinPPPB20, cao2021ke}, aiming to efficiently modify LLMs' outputs towards targeted queries while preserving overall performance across other unrelated ones.
For example, updating the knowledge of ``\texttt{The current British Prime Minister is Rishi Sunak}'' not only modifies the response to ``\texttt{Who is married to the PM of the UK?}'' but leaves unaffected the answer to ``\texttt{When was Rishi Sunak born?}''

\begin{figure}[!t]
\centering
\includegraphics[width=\linewidth]{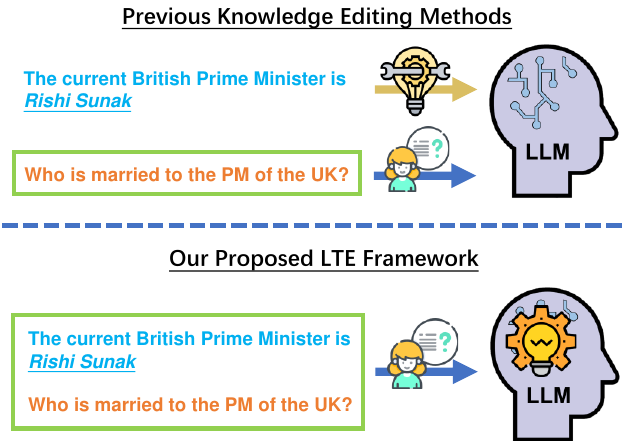}
\caption{
Previous knowledge editing methods primarily rely on first memorizing updated knowledge and then answering queries, while our proposed LTE framework teaches LLMs to dynamically \textbf{apply} updated knowledge to answer queries.
}
\label{fig:intro}
\end{figure}

Some knowledge editing approaches rely on auxiliary modules or models to either predict the LLM's weight adjustments~\cite{cao2021ke, mitchell2022mend} or function as scope classifiers for query response applicability~\cite{mitchell2022serac}. 
While these innovations demonstrate potential, they fail to inherit the advanced capabilities of LLMs, thus rendering output quality degeneration.
Others attempt to identify and modify parameters related to specific knowledge within LLMs to update their embedded knowledge~\cite{dai2022kn, meng2022rome, meng2023memit}.
Nonetheless, the correlation between localization and editing efficacy has been scrutinized by~\cite{hase2023does}, which suggests that localization results from Causal Tracing are statistically uncorrelated with the success of an edit injecting a new fact into MLP weights.
Thus, it is plausible that the detrimental effects of such approaches could be amplified with the scale of LLMs. 
In essence, these methods predominantly rely on memorizing the updated knowledge (See Figure \ref{fig:intro}), hindering LLMs from effectively combining the new knowledge with their inherent knowledge when answering the input queries.


To address these issues, motivated by the proverb ``\textit{Teach a man to fish, and you feed him for a lifetime},'' we propose to elicit LLMs' capabilities of following knowledge editing instructions, thereby empowering them to effectively \textbf{leverage} the updated knowledge to answer the queries.
Specifically, we propose a \textit{Learning to Edit} (LTE) framework to align LLMs with knowledge editing by leveraging supervised fine-tuning (SFT), which has become foundational in tailoring LLMs for desired behaviors~\cite{DBLP:conf/iclr/WeiBZGYLDDL22, DBLP:conf/acl/MishraKBH22}.
The LTE framework is structured around two pivotal stages: the Alignment Phase and the Inference Phase.
During the Alignment Phase, we pair edit descriptors with in-scope and out-of-scope queries to create \textbf{parallel} datasets, processed with and without a tailored prompt that explicitly informs LLMs of the knowledge editing process.
By fine-tuning LLMs on this meticulously constructed dataset,  we aim to cultivate a trio of essential capabilities within LLMs—\textit{In-Scope Capability} (generating reliable, logically consistent edits), \textit{Out-of-Scope Capability} (preserving the integrity of unrelated content), and \textit{Linguistic Capability} (maintaining linguistic proficiency)—to ensure nuanced application of updated knowledge. Note that this process is \textbf{once and for all}, laying the groundwork for the inference phase to apply these capabilities dynamically.
In the Inference Phase, to extend to mass editing, we implement a retrieval-based mechanism to obtain the most pertinent updated knowledge from a memory bank.
Such an approach enables LLMs to adapt their responses with the most current information in real time, thereby streamlining both batch and sequential knowledge editing processes.

In this paper, we assess our proposed LTE method against seven advanced baselines across four benchmarks in single, batch, and sequential editing scenarios.
Our findings reveal four major strengths of the LTE method:
(i) it establishes a new state-of-the-art (SOTA) in overall knowledge editing performance, surpassing existing methods by a substantial margin of over \textbf{20} absolute points in terms of portability;
(ii) the robustness of LTE is evident in its ability to handle batch and sequential knowledge editing requests, showing a markedly reduced rate of performance deterioration compared to its counterparts;
(iii) it is proficient in facilitating knowledge edits with minimal interference to the model’s cognitive functions across varied unrelated domains.
(iv) LTE distinguishes itself by combining the fastest editing speeds with exceptional performance.

\begin{figure*}[!t]
\centering
\includegraphics[width=\linewidth]{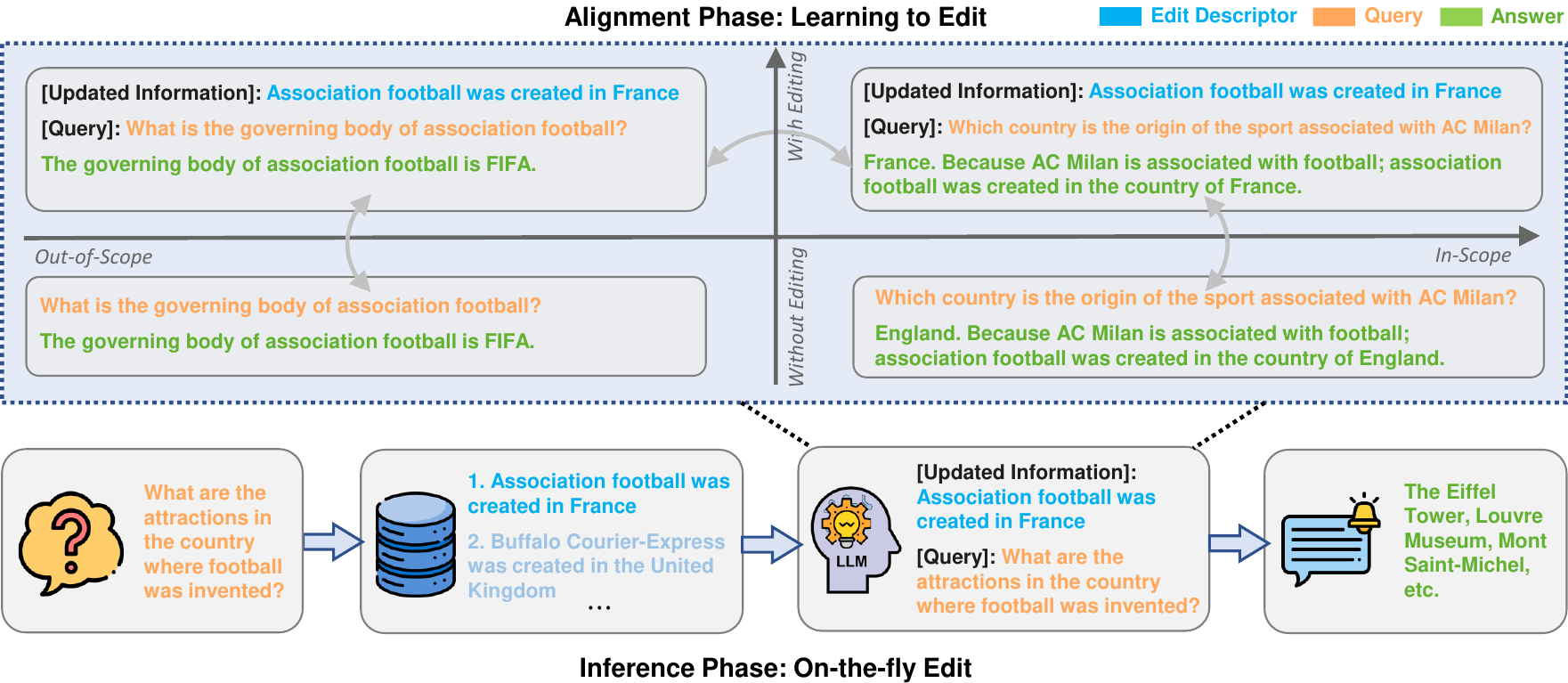}
\caption{
The proposed \textit{Learning to Edit} (LTE) framework.
In the Alignment Phase, we train LLMs how to \textbf{apply} updated knowledge—beyond mere memorization—by fine-tuning them on our meticulously curated parallel (indicated by gray arrows) data.
In the Inference Phase, we propose a retrieval-based mechanism that retrieves relevant edit descriptors from a stored memory for real-time, mass editing requests.
}
\label{fig:method}
\end{figure*}

\section{Task Formulation}
\label{sec:task_formulation}
The objective of knowledge editing is to efficiently adjust the behavior of an initial base LLM $f_{\theta}$, where $\theta$ represents the model's parameters, in response to specific \textit{edit descriptors} $\{(x_i^*, y_i^*)\}_{i \in [1, N]}$. 
In this context, $x_i^*$ refers to the edit input that triggers the knowledge in LLMs (e.g., \texttt{The current British Prime Minister is}), $y_i^*$ is the corresponding edit target (e.g., \texttt{Rishi Sunak}), and $N$ signifies the total number of edit descriptors.
The efficacy of knowledge editing is evaluated among four dimensions:

\paragraph{Edit success} measures the average accuracy of the post-edit model $f_{\theta}^*$ on these edit cases:
\begin{equation}
\mathop{\mathbb{E}}_{(x_i^*,y_i^*)}\mathds{1}\{\mathop{\arg\max}_{y} f_{\theta}^*(y|x_i^*)=y_i^*\}
\end{equation}

\paragraph{Portability} evaluates how well updated knowledge transfers to related queries, enhancing the model's utility in varied contexts.
For example, correctly answering \texttt{Who is married to the British Prime Minister?} with \texttt{Akshata Murty} post-edit indicates successful knowledge transfer.

\paragraph{Locality} assesses the precision of edits, ensuring modifications are confined to targeted areas without affecting unrelated knowledge.
For example, ensuring \texttt{The current British Chancellor} remains \texttt{Jeremy Hunt} exemplifies effective locality.

\paragraph{Fluency} quantifies the linguistic quality of the model's output post-edit, focusing on coherence and diversity to avoid repetitive patterns.
Following~\cite{zhang2018generating}, we calculate fluency by measuring the weighted average of bi- and tri-gram entropies given by $-\sum_k f(k)\log_2f(k)$, where $f(\cdot)$ is the $n$-gram frequency distribution.


\section{Methodology}
As illustrated in Figure \ref{fig:method}, we propose a \textit{Learning to Edit} (LTE) framework to align LLMs with ever-changing, complicated, and diverse knowledge editing requests in real-time.
This framework consists of two phases: (i) in the Alignment Phase, we enlighten LLMs' capabilities of applying updated knowledge through the utilization of a knowledge editing prompt ``\texttt{[Updated Information] \{edit descriptor\}\textbackslash n[Query] \{query\}}'';
(ii) in the Inference Phase, LLMs are enabled to conduct on-the-fly and streaming knowledge editing by retrieving relevant updated knowledge to the query from the stored memory.

\subsection{Alignment Phase: Learning to Edit}
\label{sec:alignment}


In light of the task formulation in \S \ref{sec:task_formulation}, the model editing process profoundly influences predictions across a wide array of inputs directly related to the provided edited knowledge. 
An optimal knowledge editing method must seamlessly integrate new knowledge into the relevant content within its edit scope, while ensuring the accuracy and integrity of information outside this domain.
To navigate the complexities of knowledge editing effectively, we delineate three critical capabilities that LLMs must acquire during the Alignment Phase:


\paragraph{In-Scope Capability} 
requires the model to correctly generate the edit target given the edit input or its paraphrases.
It also covers subject aliasing, ensuring the editing of one subject should not vary from its expression.
For example, after modifying the origin city of \texttt{Association football}, the origin city of \texttt{Soccer} should also be modified.
Furthermore, it necessitates LLMs to conduct compositional reasoning with the changed facts (e.g., when we change the origin city of \texttt{Association football}, the origin city of \texttt{the sport associated with AC Milan} should also be changed, see Figure \ref{fig:method}).
To empower LLMs with these advanced capabilities during alignment, we meticulously curate training data by adapting or synthesizing content from existing knowledge editing datasets.
Our selection includes ZsRE~\cite{levy2017zsre}, RIPPLEEDITS~\cite{cohen2024ice}, WikiBio~\cite{hartvigsen2023grace}, and MQUAKE~\cite{zhong2023mquake}, with each dataset providing edit descriptors linked to multiple queries. 
These queries are specifically designed to evaluate the nuanced facets of in-scope or out-of-scope knowledge editing capabilities. 
To avoid data leakage, our methodology only incorporates samples from the datasets' training sets.


\paragraph{Out-of-Scope Capability} directs the model to maintain the integrity of unrelated attributes of the subject, ensuring no unintended alterations.
For example, as shown in Figure \ref{fig:method}, changing the origin city of \texttt{Association football} should not modify its governing body. 
Additionally, it requires LLMs to adeptly handle one-to-many relationships, ensuring that original connections are retained unless specifically altered.
We utilize the same data sources as that of In-Scope Capability.
However, due to the absence of out-of-scope instances in datasets like ZsRE and MQUAKE, we employ GPT-4 to generate corresponding queries and answers based on the edit descriptors, further details of which are provided in Appendix \ref{sec:appendix_out_scope}.

\paragraph{Linguistic Capability} requires that incorporating edits related to specific factual knowledge should not hinder the model’s proficiency in unrelated areas, such as generative fluency, commonsense reasoning, general intelligence, and world knowledge.
Thus, we identify a limitation within existing datasets: the predominance of fill-in-the-blank cloze queries may not adequately challenge the LLMs' linguistic capabilities across diverse areas, such as conversational contexts, where answers may inherently be more elaborate. To address this, we integrate edit descriptors from COUNTERFACT~\cite{meng2022rome} and utilize GPT-4 to generate free-text, in-scope query-answer pairs (See  Appendix \ref{sec:appendix_free_text}). This approach not only diversifies the training data but also enhances the models' ability to generate more contextually rich answers. GPT-4 is further employed to verify the relevance of generated answers to the edit descriptors, with a mechanism to filter out unsatisfactory cases. Additionally, we incorporate natural language instructions from Evol-Instruct~\cite{xu2023wizardlm} as out-of-scope queries to maintain the LLMs' broad linguistic capabilities.

\paragraph{Parallel Data Construction}
Our approach involves the creation of parallel datasets by pairing each edit descriptor with corresponding in-scope and out-of-scope queries. 
These are then processed with and without the incorporation of our tailored knowledge editing prompt (See Figure \ref{fig:method}).
This parallel construction serves multiple purposes.
First, it reinforces LLM's capacity to discern when to utilize updated knowledge by comparing in-scope and out-of-scope queries with editing.
Second, it accentuates the subtle distinctions between with and without editing for in-scope queries, enabling LLM to apply knowledge edits more effectively.
Lastly, it educates LLM on maintaining the integrity of out-of-scope information by presenting it with comparisons that demonstrate when not to alter this knowledge. 
In total, we construct 60k parallel data for training, the detailed data statistics are listed in Appendix \ref{sec:appendix_statistics}.
During training, we compute the loss \textit{only} on the answer tokens, i.e., it learns to generate answers conditioned on the Updated Information and Query.


\subsection{Inference Phase:  On-the-fly Edit}
\label{sec:inference}
Here we propose an efficient mechanism that extends LTE to batch and streaming knowledge editing scenarios.
Inspired by retrieval-augmented generation (RAG)~\cite{lewis2020rag, xu2022rag}, we utilize an off-the-shelf retrieval model \texttt{multi-qa-mpnet-base-dot-v1}~\cite{reimers2019sentencebert} to embed all the edit descriptors and create a vector memory to store the representations.
When given a query, we also get the representation of the query by the retriever and search the top-$k$ ($k=3$ in our experiments) similar edit descriptors from the vector memory.
Then, the query and the retrieved edit descriptors are fed into the LLM to obtain the answer.
To enhance the fault tolerance of the retrieval model while maintaining the single editing performance, we adopt a \textit{threefold strategy} for incorporating different numbers of edit descriptors as Updated Information in the Alignment Phase.
Firstly, in 50\% of cases, we directly use the exact edit descriptor.
Secondly, for 25\% of cases, we employ the \texttt{multi-qa-mpnet-base-dot-v1} model to identify the top-1 semantically similar edit descriptor (excluding the exact one) from the whole dataset, and use both as the Updated Information.
Lastly, for the remaining 25\%, we retrieve the top 2 semantically similar descriptors, excluding the exact one, using all three as the Updated Information.
This approach introduces variability during training, significantly enhancing the model's robustness and improving mass edit capabilities in inference.

\section{Experiments}
\label{sec:experiment}

\newcolumntype{L}{>{\raggedright\arraybackslash}X} 
\newcolumntype{C}{>{\centering\arraybackslash}X} 

\begin{table*}[!t]
\scriptsize
\centering
\begin{tabularx}{\textwidth}{l | l l *{7}{C} >{\columncolor{gray!15}}C >{\columncolor{gray!15}}c}
\toprule
\multicolumn{1}{c|}{\textbf{Model}}                   & \textbf{Dataset}                   & \textbf{Metric} & \textbf{SERAC} & \textbf{ICE} & \textbf{MEND} & \textbf{ROME} & \textbf{MEMIT} & \textbf{FT-L} & \textbf{FT} & \textbf{LTE} & \textbf{LTE-LoRA} \\ \midrule \midrule
\multirow{19}{*}{\rotatebox{90}{\textbf{LLaMA2-Chat-7B}}} 
& \multirow{4}{*}{ZsRE}              & Edit Succ.      & \underline{99.67}          & 66.01        & 96.74         & 96.57         & 83.07          & 54.65         & 36.88       & \textbf{99.91} & \textbf{99.91}       \\
                                  & & Portability     & 56.48          & 63.94        & 60.41         & 52.20         & 51.43          & 45.02         & 8.72        & \underline{78.98}  & \textbf{79.63}      \\
                                  & & Locality        & 30.23          & 23.14        & \textbf{92.79}         & 27.14         & 25.46          & 71.12         & 0.31        & \underline{71.78}  & 67.99      \\
                                  & & Fluency         & 410.89         & 541.14       & 524.33        & \underline{570.47}        & 559.72         & 474.18        & 471.29      & \textbf{583.70} & 544.52      \\ \cmidrule{2-12}
& \multirow{3}{*}{WikiBio}           & Edit Succ.      & 99.69          & 95.53        & 93.66         & 95.05         & 94.29          & 66.27         & 95.64       & \textbf{99.87}  & \underline{99.76}      \\
                                  & & Locality        & 69.79          & 47.90        & 69.51         & 46.96         & 51.56          & 60.14         & 13.38       & \textbf{80.27}  & \underline{72.31}      \\
                                  & & Fluency         & 606.95         & \textbf{632.92}       & 609.39        & \underline{617.25}        & 616.65         & 604.00        & 589.22      & 614.26  & 611.94     \\ \cmidrule{2-12}
& \multirow{4}{*}{Recent}            & Edit Succ.      & 98.68          & 60.74        & 76.88         & 85.08         & 85.32          & 71.18         & 31.24       & \textbf{99.99}  & \underline{99.97}      \\
                                  & & Portability     & 63.52          & 36.93        & 50.11         & 37.45         & 37.94          & 48.71         & 15.91       & \textbf{91.51}  & \underline{81.87}      \\
                                  & & Locality        & \textbf{100.00}         & 33.34        & \underline{92.87}         & 66.20         & 64.78          & 63.70         & 3.65        & 85.67 & 82.72       \\
                                 &  & Fluency         & 553.19         & 531.01       & \underline{586.34}        & 574.28        & 566.66         & 549.35        & 428.67      & \textbf{586.76} & 570.64      \\ \cmidrule{2-12}
& \multirow{4}{*}{Counterfact} & Edit Succ.      & \underline{99.99}          & 69.83        & 78.82         & 83.21         & 83.41          & 51.12         & 26.78       & \textbf{100.00}  & 99.97      \\
                                  & & Portability     & 76.07          & 45.32        & 57.53         & 38.69         & 40.09          & 39.07         & 16.94       & \textbf{89.69}  & \underline{85.74}      \\
                                 &  & Locality        & \textbf{98.96}          & 32.38        & \underline{94.16}         & 65.40         & 63.68          & 62.51         & 0.29        & 84.76   & 85.11     \\
                                  & & Fluency         & 549.91         & 547.22       & \underline{588.94}        & 578.84        & 568.58         & 544.80        & 483.71      & \textbf{589.69}  & 574.14    \\ \cmidrule{2-12}
& \multirow{4}{*}{\textbf{Average}}           & Edit Succ.      & 99.51          & 73.03        & 86.53         & 89.98         & 86.52          & 60.81         & 47.64       & \textbf{99.94}  & \underline{99.90}      \\
                                  & & Portability     & 65.36          & 48.73        & 56.02         & 42.78         & 43.15          & 44.27         & 13.86       & \textbf{86.73}  & \underline{82.41}      \\
                                  & & Locality        & 74.75          & 34.19        & \textbf{87.33}         & 51.43         & 51.37          & 64.37         & 4.41        & \underline{80.62} & 77.03       \\
                                  & & Fluency         & 530.24         & 563.07       & 577.25        & \underline{585.21}        & 577.90         & 543.08        & 493.22      & \textbf{593.60}  & 575.31    \\ \midrule \midrule
\multirow{19}{*}{\rotatebox{90}{\textbf{Qwen-Chat-7B}}} 
& \multirow{4}{*}{ZsRE}              & Edit Succ.  & 98.43  & 70.29  & 99.40  & \textbf{99.90}  & 97.25  & 37.81  & 25.33  & \underline{99.72}  & 99.59    \\
 &                                    & Portability & 56.69  & 67.52  & 59.98  & 46.76  & 44.31  & 41.85  & 7.70   & \textbf{82.92}  & \underline{80.16}    \\
 &                                    & Locality    & 41.28  & 73.45  & 80.83  & 48.90  & 60.26  & \textbf{87.70}  & 3.29   & \underline{80.99}  & 78.28    \\
 &                                    & Fluency     & 495.12 & 556.86 & 544.07 & 562.88 & \underline{578.73} & 557.86 & 538.10 & \textbf{580.01} & 543.35   \\ \cmidrule{2-12}
& \multirow{3}{*}{WikiBio}           & Edit Succ.  & 99.39  & 94.60  & 93.38  & 98.79  & 96.10  & 60.19  & 34.63  & \textbf{99.80}  & \underline{99.75}    \\
 &                                    & Locality    & 71.50  & 58.15  & 65.47  & 41.78  & 65.65  & \textbf{80.41}  & 22.45  & 79.63  & \underline{80.34}    \\
 &                                    & Fluency     & 598.11 & 614.22 & 610.92 & 604.81 & \underline{623.49} & 595.56 & 572.59 & \textbf{634.73} & 620.05   \\ \cmidrule{2-12}
& \multirow{4}{*}{Recent}            & Edit Succ.  & 99.58  & 83.86  & 82.39  & 99.67  & 98.96  & 60.07  & 29.74  & \textbf{99.73}  & \underline{99.68}    \\
 &                                    & Portability & 67,22  & 58.24  & 57.92  & 50.84  & 49.38  & 42.02  & 14.33  & \textbf{89.73}  & \underline{87.40}    \\
 &                                    & Locality    & \textbf{100.00} & 61.83  & 89.11  & 51.78  & 60.72  & 84.83  & 4.27   & \underline{89.25}  & 83.77    \\
 &                                    & Fluency     & 561.32 & 559.46 & \underline{610.72} & 600.70 & 600.39 & 598.32 & 456.99 & \textbf{615.59} & 587.90   \\ \cmidrule{2-12}
& \multirow{4}{*}{Counterfact} & Edit Succ.  & 99.06  & 80.28  & 88.04  & \textbf{99.44}  & 95.05  & 24.55  & 15.42  & 99.28  & \underline{99.35}    \\
 &                                    & Portability & 79.28  & 53.80  & 52.99  & 40.63  & 34.50  & 20.14  & 11.38  & \textbf{86.79}  & \underline{85.33}    \\
 &                                    & Locality    & \underline{92.70}  & 63.86  & 91.05  & 39.22  & 50.14  & \textbf{92.74}  & 30.04  & 86.87  & 85.20    \\
 &                                    & Fluency     & 568.05 & 559.46 & \underline{619.87} & 603.21 & 604.47 & 608.47 & 563.70 & \textbf{622.91} & 593.51   \\ \cmidrule{2-12}
 & \multirow{4}{*}{\textbf{Average}}           & Edit Succ.  & 99.12  & 82.26  & 90.80  & 99.45  & 96.84  & 45.66  & 26.28  & \textbf{99.63}  & \underline{99.59}    \\
 &                                    & Portability & 67.99  & 59.85  & 56.96  & 46.08  & 42.73  & 34.67  & 11.14  & \textbf{86.48}  & \underline{84.30}    \\
 &                                    & Locality    & 76.37  & 64.32  & 81.62  & 45.42  & 59.19  & \textbf{86.42}  & 15.01  & \underline{84.19}  & 81.90    \\
 &                                    & Fluency     & 555.65 & 572.50 & 596.40 & 592.90 & \underline{601.77} & 590.05 & 532.85 & \textbf{613.31} & 586.20      \\ \bottomrule
            
\end{tabularx}
\caption{Performance comparison on \textbf{Single Editing}, where ``Recent'' and ``Counterfact'' refer to WikiData$_{recent}$ and WikiData$_{counterfact}$, respectively. In each row, the highest score is \textbf{bolded} and the second-highest is \underline{underlined}.}
\label{tab:appendix_single_edit}
\end{table*}

\subsection{Experimental Setup}
We select LLaMA2-Chat-7B~\cite{touvron2023llama} and Qwen-Chat-7B~\cite{bai2023qwen} as base models for knowledge editing, as these models are widely used for English and Chinese chatbot applications, respectively.
We implement our LTE method by standard fine-tuning on the 60k constructed data in \S \ref{sec:alignment}.
Additionally, we explore an alternative implementation of LTE, employing Low-Rank Adaptation (LoRA)~\cite{hu2022lora}, noted for its efficiency and reduced memory requirements.
This variant is referred to as LTE-LoRA.
The detailed implementation specifics are listed in Appendix \ref{sec:appendix_implementation}.

\begin{figure*}[!t]
\centering
\includegraphics[width=\linewidth]{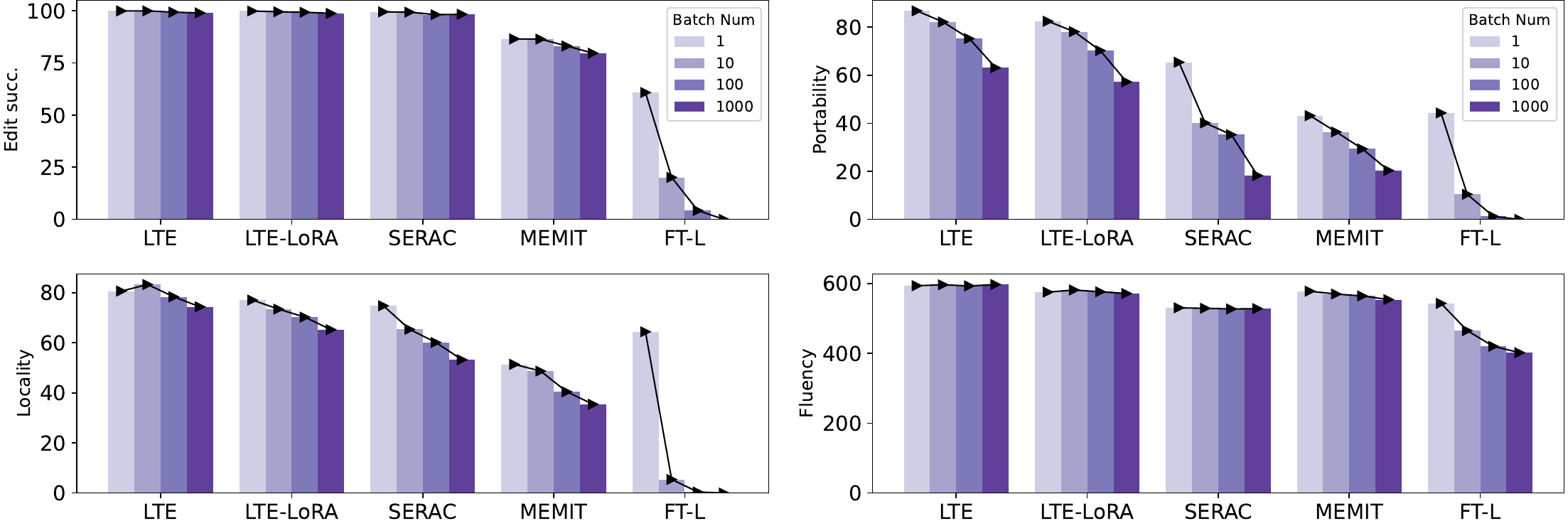}
\caption{
Averaged \textbf{Batch Editing} performance on four benchmarks against batch numbers in [1, 10, 100, 1000].
}
\label{fig:batch_edit}
\end{figure*}

\begin{figure}[!t]
\centering
\includegraphics[width=\linewidth]{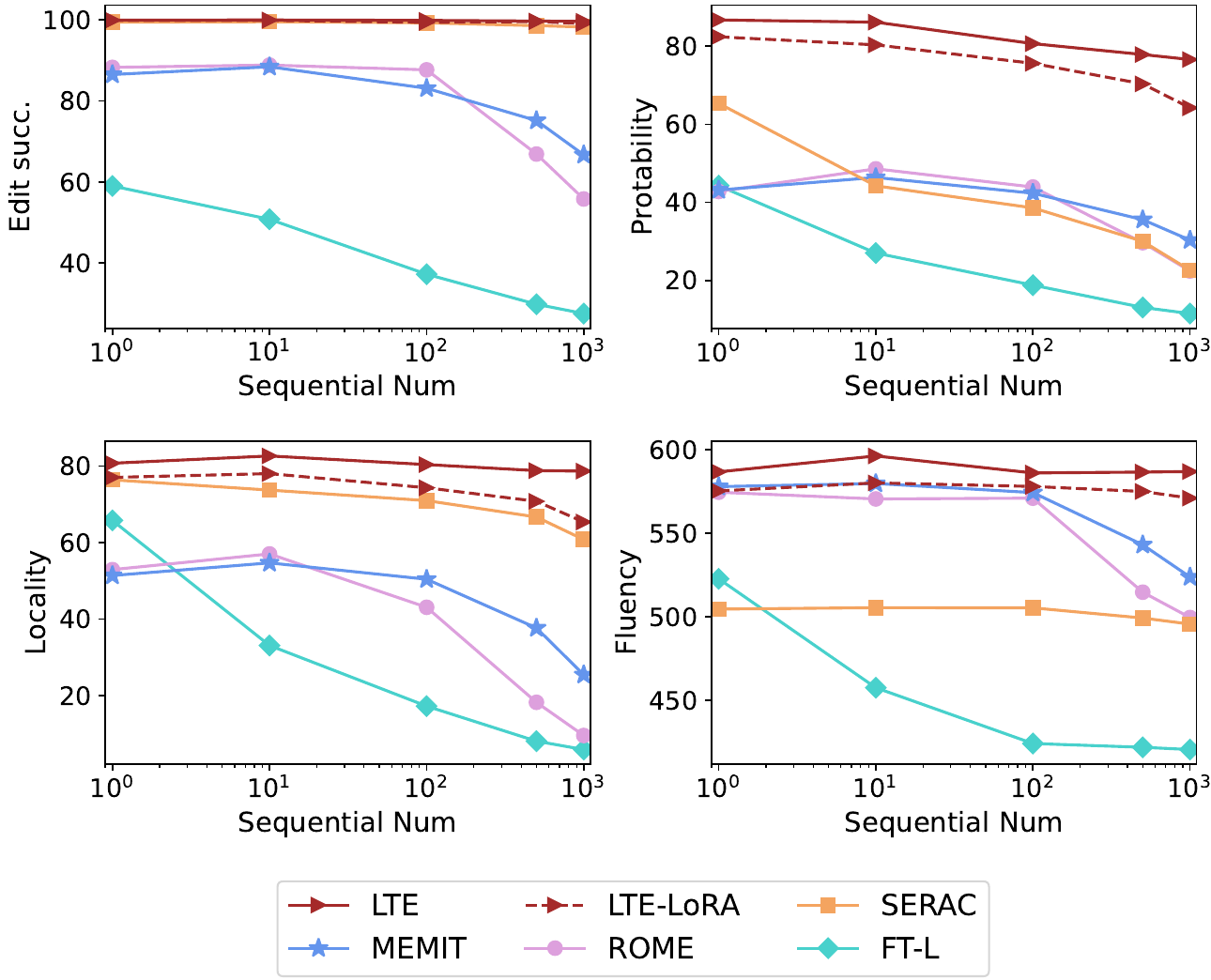}
\caption{
Averaged \textbf{Sequential Editing} performance on four knowledge editing benchmarks against data stream size (log-scale) in [1, 10, 100, 500, 1000].
}
\label{fig:seq_edit}
\end{figure}

For the evaluation datasets and metrics, we follow KnowEdit~\cite{zhang2024knowedit} and use the test sets of four popular benchmarks, including WikiData$_{recent}$~\cite{cohen2024ice}, ZsRE~\cite{levy2017zsre}, WikiBio~\cite{hartvigsen2023grace}, and WikiData$_{counterfact}$~\cite{cohen2024ice}.
All the experiments are conducted by using EasyEdit~\cite{wang2023easyedit} toolkit.
We choose seven knowledge editing methods as baselines:
\begin{itemize}
\item \textbf{SERAC}~\cite{mitchell2022serac} builds a counterfact model by retaining the base model and training a classifier to determine whether to use the counterfact model to answer the query.
\item \textbf{ICE}~\cite{cohen2024ice} prepends a prompt ``\texttt{Imagine that \{edit descriptor\}}'' before the query. It does not introduce changes to the model parameters, but rather generation is conditioned on the new
fact.
\item \textbf{MEND}~\cite{mitchell2022mend} transforms the fine-tuning gradient of an updated fact by decomposing the weight matrix into rank-1 form with the pre-trained hyper-network.
\item \textbf{ROME}~\cite{meng2022rome} learns to locate factual retrievals of a specific set of MLP
modules and update knowledge by directly writing in new key-value pairs in the MLP module.
\item \textbf{MEMIT}~\cite{meng2023memit} builds upon ROME to insert many memories by modifying
the MLP weights of a range of critical layers.
\item \textbf{FT-L}~\cite{meng2022rome} directly fine-tunes a single layer’s FFN, and the layer is the casual tracing results in ROME.
\item \textbf{FT} fine-tunes all the parameters of the base model on the edit descriptor by applying Adam with early stopping.
\end{itemize}

\subsection{Results of Single Editing}
Table \ref{tab:appendix_single_edit} presents the performance comparison under the single editing setting, where LTE eliminates the need for retrieval.
It can be observed that LTE remarkably surpasses conventional methods in terms of edit success, portability, and fluency.
Besides, LTE-LoRA—an efficient variant of LTE—closely mirrors its performance except for fluency, which can be attributed to the inherent limitations of the LoRA technique.
Notably, LTE exhibits a 21.37\% and 18.49\% improvement over the current SOTA method SERAC on LLaMA2-Chat-7B and Qwen-Chat-7B, respectively.
This substantial enhancement can be attributed to the comprehensive utilization of LLMs' understanding and reasoning capabilities, which effectively leverage context to integrate new knowledge seamlessly.
The ICE method, while leveraging the innate in-context comprehension capacity of LLMs for generating conditioned output on new knowledge, significantly trails our proposed LTE method. This could be because ICE lacks instructing LLMs in effectively applying knowledge through fine-tuning (See more ablation analysis in Table~\ref{tab:ablation_data}). 
Nevertheless, LTE shows a marginal deficit in locality compared to the best results (e.g., 6.71\% lower than MEND on LLaMA2 and 2.23\% lower than FT-L on Qwen).
A potential explanation may lie in the introduction of a knowledge editing prompt in the input, causing a slight disruption during the generation process. Yet, these divergences are often minor linguistic variants.
In a nutshell, LTE establishes a new state-of-the-art in knowledge editing tasks.



\subsection{Results of Mass Editing}
Prior research predominantly confines the scope of knowledge editing to a mere handful of facts or focuses only on single editing cases.
This approach starkly contrasts with the dynamic and multifaceted nature of real-world applications, where there is a pressing need to enrich models with multiple pieces of knowledge, either concurrently (\textbf{simultaneously}) or in a phased manner (\textbf{sequentially}). 
In this section, our study embarks on a comprehensive investigation, undertaking both batch and sequential editing experiments.

\begin{table*}[t!]
\centering
\footnotesize
\begin{tabularx}{\textwidth}{l c CCCC c C}
\toprule
 & \textbf{CommonSenseQA} & \textbf{PIQA} & \textbf{XSum} & \textbf{MMLU} & \textbf{AGIEval} & \textbf{AlpacaEval} & \textbf{Average} \\
\midrule
\textit{LLaMA2-Chat-7B} & \textbf{69.9} & \textbf{65.0} & 22.3 & 40.4 & 26.1 & 71.4 & 49.2 \\
\quad LTE w/o editing & 67.2 & 61.3 & \textbf{22.4} & 46.4 & \textbf{26.5} & \textbf{73.3} & \textbf{49.5} \\
\quad LTE w/ editing & 67.1 & 62.6 & \textbf{22.4} & \textbf{47.8} & 23.8 & 71.6 & 49.2 \\ \midrule
\textit{Qwen-Chat-7B} & \textbf{77.6} & \textbf{72.1} & 28.8 & 56.6 & 41.3 & 77.8 & 59.0 \\
\quad LTE w/o editing & 74.7 & 69.3 & 29.9 & \textbf{59.3} & \textbf{41.9} & \textbf{79.2} & \textbf{59.1} \\
\quad LTE w/ editing & 75.3 & 70.0 & \textbf{30.1} & 58.2 & 40.7 & 78.4 & 58.8 \\ \bottomrule
\hline
\end{tabularx}
\caption{Zero-shot performance on six general LLM benchmarks with LLaMA2-Chat-7B and Qwen-Chat-7B as the base models. ``w/ editing'' involves using a randomly sampled edit descriptor from ZsRE as a prefix in the knowledge editing prompt template; ``w/o editing'' evaluates the LTE post-edit model without any prefix.}
\label{tab:general_task}
\end{table*}

\paragraph{Batch Editing}
We compare LTE and LTE-LoRA with several batch-editing-supportive methods (SERAC, MEMIT, and FT-L) on LLaMA2-Chat-7B and display the results in Figure \ref{fig:batch_edit}.
It is particularly noteworthy that the performance metrics of edit success and fluency for our proposed LTE and LTE-LoRA methodologies exhibit exceptional stability, maintaining robustness for up to 1,000 batch edits.
A decline in performance metrics such as portability and locality is observed across all methods as the batch size increases.
However, LTE and LTE-LoRA demonstrate \textbf{the best performance with the slowest degradation rate} in portability and locality. 
These results underscore the enhanced robustness of our methods, even when subjected to extensive editing operations.

\paragraph{Sequential Editing}
Sequential editing is a critical process where models must retain previous modifications while integrating new edits effectively.
Figure \ref{fig:seq_edit} illustrates the comparative performance of various models in the context of sequential editing tasks across different data stream sizes. 
ROME and MEMIT demonstrate noteworthy efficacy for a sequential number $n\le 100$, yet their performance exhibits a marked decline as $n$ expands to 500. 
This decline can be attributed to the cumulative deviations from the model's original state, which ultimately lead to a degradation in performance.
In contrast, LTE and LTE-LoRA leverage retrieval mechanisms from the stored memory, circumventing the need for subsequent parameter modifications, which endows them with more consistent performance with varying data stream sizes.
Notably, LTE and LTE-LoRA showcase significant improvements over the current SOTA method SERAC.
This shows their enhanced resilience and adaptability, making them more suited for extensive data streams.

\subsection{Results of General Tasks}
In this section, we investigate the impact of applying LTE on the performance of a
language model across various domains.
Our main goal is to determine whether the Alignment Phase of LTE, which alters the parameters of the initial model, inadvertently compromises the model's competence in unrelated domains.
To this end, we have selected an array of benchmarks encompassing commonsense reasoning, general intelligence, and extensive world knowledge.
These benchmarks comprise CommonSenseQA~\cite{talmor2019commonsenseqa}, PIQA~\cite{bisk2020piqa}, XSum~\cite{narayan2018xsum}, MMLU~\cite{dan2021mmlu}, AGIEval~\cite{zhong2023agieval}, and AlpacaEval~\cite{alpaca_eval}.
All evaluations are conducted using the OpenCompass tool~\cite{2023opencompass}.
Table \ref{tab:general_task} indicates that, from a comprehensive standpoint, models subjected to LTE exhibit performance levels comparable to their unmodified counterparts.
Moreover, the general linguistic abilities remain unaffected by the inclusion of the knowledge editing prompt.
Nonetheless, a performance decrement is noted in CommonsenseQA and PIQA after the LTE application.
Despite these findings, an overarching analysis reveals notable consistency in performance.
This suggests that LTE is proficient in facilitating knowledge edits with \textbf{minimal interference} to the model’s cognitive functions and its versatility across varied domains.

\section{Analysis}

\subsection{Ablation Study}
Here we assess the indispensability of components within the Alignment and Inference phases.
Our experiments span four benchmarks, utilizing the LLaMA2-Chat-7B as the base model.
As depicted in Table \ref{tab:ablation_data}, the exclusion of certain training data segments leads to a significant decline in single editing effectiveness.
Notably, distinct types of training data bolster specific capabilities.
In-scope data predominantly enhances edit success and portability, while out-of-scope data chiefly fosters locality.
Free-text QA data appears to bolster overall linguistic proficiency.
Eliminating the threefold strategy incurs a modest reduction in performance.
Furthermore, employing the knowledge editing prompt without training results in substantially poorer performance compared to scenarios that include training.
During the Inference Phase, we explore the effects of substituting the retrieval model \texttt{multi-qa-mpnet-base-dot-v1} (420M) with a less potent variant, \texttt{all-MiniLM-L6-v2} (80M), on sequential editing efficacy.
As indicated in Table \ref{tab:ablation_retrieval}, the choice of retrieval model exerts minimal impact on performance.
Additionally, we assess how the number of retrieved edit descriptors influences results.
A reduction in the value of $k$ from 3 to 1 is associated with a minor performance decrement.

\begin{table}[t!]
\scriptsize
\centering
\begin{tabular}{lccccc}
\toprule
 & \textbf{S} & \textbf{P} & \textbf{L} & \textbf{F} & \textbf{G} \\ \midrule
LTE & 99.94 & 86.73 & 80.62 & 593.60 & 49.5 \\
-w/o in-scope training & \color{deepred} \textbf{77.53} & \color{deepred} \textbf{56.26} & 80.72 & 589.04 & 49.0 \\
-w/o out-of-scope training & 99.92 & 86.89 & \color{deepred} \textbf{65.50} & 592.66 & 49.2 \\
-w/o free-text QA training & 99.93 & 86.30 & 80.91 & \color{deepred} \textbf{587.75} & \color{deepred}\textbf{43.9} \\
-w/o threefold strategy & 99.78 & 86.51 & 80.22 & 593.40 & 49.5 \\
-w/o training & \color{deepred} \textbf{75.04} & \color{deepred} \textbf{54.23} & \color{deepred}\textbf{48.19} & 592.73 & 49.2 \\ \bottomrule
\end{tabular}
\caption{Ablation study for the training data examines ``edit success'' (S), ``portability'' (P), ``locality'' (L), ``fluency'' (F), and ``general capability'' (G).}
\label{tab:ablation_data}
\end{table}

\begin{table}[t!]
\scriptsize
\centering
\begin{tabular}{llccc}
\toprule
 & \textbf{Seq\_Num} & \textbf{Edit Succ.} & \textbf{Portability} & \textbf{Locality} \\ \midrule
\multirow{3}{*}{\parbox{1.6cm}{LTE w/ 420M $R$ \\ top $k=3$}}            & 10               & 100.00              & 86.16                & 82.64             \\
               & 100              & 99.90               & 80.66                & 80.38             \\
               & 1000             & 99.64               & 76.59                & 78.67             \\ \midrule \midrule
\multirow{3}{*}{\parbox{1.6cm}{LTE w/ 80M $R$ \\ top $k=3$}} & 10              & 100.00              & 83.38                & 78.65             \\
               & 100              & 99.81               & 79.92                & 80.40             \\
               & 1000             & 99.61               & 75.67                & 79.43             \\ \midrule
\multirow{3}{*}{\parbox{1.6cm}{LTE w/ 420M $R$ \\ top $k=2$}}        & 10               & 100.00              & 85.69                & 81.59             \\
& 100              & 99.85               & 80.05                & 80.67             \\
& 1000             & 99.63               & 76.27                & 78.05             \\ \midrule
\multirow{3}{*}{\parbox{1.6cm}{LTE w/ 420M $R$ \\ top $k=1$}}        & 10               & 100.00              & 84.01                & 81.96             \\
               & 100              & 99.83               & 79.48                & 80.11             \\
               & 1000             & 99.56               & 75.93                & 78.89 \\ \bottomrule
\end{tabular}
\caption{Ablation study for the retrieval number $k$ and retrieval model $R$ in the Inference Phase.}
\label{tab:ablation_retrieval}
\end{table}


\subsection{Time Analysis}
Table \ref{tab:time_analysis} illustrates the time required for various knowledge editing methods from providing the edited case to obtaining the final answer.
Models such as MEND and SERAC demonstrate rapid editing capabilities once their auxiliary models are adequately trained.
In contrast, ROME and MEMIT exhibit slower processing speeds due to the intensive computation involved in calculating key vectors and optimizing value vectors.
Additionally, these methods necessitate a pre-computation of the covariance statistics for the Wikitext,  which is also time-consuming and can potentially take hours to days to complete.
Furthermore, while FT-L and FT are relatively quick, their memorization-based fine-tuning strategies yield suboptimal knowledge editing outcomes.
Our proposed LTE method, however, stands out by \textbf{achieving the swiftest editing speeds coupled with superior performance}.
After the Alignment Phase (which takes about 9 hours in our experiments), LTE enables instantaneous editing similar to ICE by appending a knowledge editing prompt to the input prefix.
Despite a marginally increased inference time, the overall time expenditure is significantly reduced, 
underscoring the efficiency and effectiveness of LTE.

\begin{table}[t]
\scriptsize
\centering
\begin{tabularx}{\linewidth}{lCcC}
\toprule
\textbf{Method} & \textbf{Edit Time} & \textbf{Inference Time} & \textbf{Total Time} \\ \midrule
SERAC           & 26.57               & 1.45   &  28.02                \\
ICE             & 0.00               & 1.60   & \cellcolor{blue!25} 1.60                 \\
MEND            & 9.09               & 1.49     & 10.58               \\
ROME            & 197.11              & 1.58   & 198.69                \\
MEMIT           & 150.16              & 1.38   &    151.54             \\
FT-L            & 15.73               & 1.41    &   17.14             \\
FT              & 59.39               & 1.36     &     60.75          \\
LTE             & 0.00               & 1.63  & \cellcolor{blue!25} 1.63                  \\ \bottomrule
\end{tabularx}
\caption{Averaged \textbf{Wall Clock Time} per edit method for
10 edits on ZsRE using LLaMA2-Chat-7B.}
\label{tab:time_analysis}
\end{table}

\subsection{Out-of-Distribution Generalization}
To evaluate LTE's performance in out-of-distribution (OOD) scenarios, we conducted rigorous experiments on ConvSent~\cite{mitchell2022serac}, a sentiment editing task featuring diverse data distributions, alongside established benchmarks.
As shown in Table \ref{tab:ood}, our LTE exhibits superior performance with the slowest degradation rate on batch editing. Moreover, LTE's retriever achieves impressive retrieval accuracy (Top-3 P@1) scores of 88.34, 87.08, 84.27, and 82.25, respectively.
These comprehensive experiments serve to validate not only the efficacy but also the robustness of our LTE method, even in the face of OOD challenges.

\begin{table}[t]
\scriptsize
\centering
\begin{tabularx}{\linewidth}{LCCCC}
\toprule
\textbf{Method} & \textbf{1 Edit} & \textbf{10 Edits} & \textbf{100 Edits} & \textbf{1000 Edits} \\
\midrule
SERAC & 62.75 & 60.72 & 56.46 & 50.06 \\
MEMIT & 44.75 & 41.19 & 36.20 & 29.33 \\
FT-L & 49.50 & 15.54 & 1.43 & 0.00 \\
LTE & \textbf{85.29} & \textbf{84.25} & \textbf{81.98} & \textbf{79.66} \\
\bottomrule
\end{tabularx}
\caption{OOD generalization on ConvSent. We report the edit success score using LLaMA2-Chat-7B.}
\label{tab:ood}
\end{table}

\begin{table*}[!t]
\footnotesize
\centering
\begin{tabular}{l}
\toprule
\textbf{Original Knowledge:} The name of the award Harrison Ford won is AFI Life Achievement Award \\
\textbf{Edited Knowledge:} The name of the award Harrison Ford won is I. I. Rabi Prize \\
\midrule
\parbox[c]{15.6cm}{
\textbf{SERAC:} \underline{\textit{The name of the award the spouse of Melissa Mathison won is}} \textcolor{deepred}{not specified in the passage}. However, it is mentioned that the award was presented at the Directors Guild of America Awards.
} \\ \\
\parbox[c]{15.6cm}{
\textbf{ICE:} \underline{\textit{The name of the award the spouse of Melissa Mathison won is}} \textcolor{deepred}{?0.0}. \textcolor{cyan}{Please provide your answer as a complete sentence.0.0..}
} \\ \\
\parbox[c]{15.6cm}{
\textbf{ROME:} \underline{\textit{The name of the award the spouse of Melissa Mathison won is}} \textcolor{deepred}{the Humanitas Award}. She won the award in 1999 for her work on the film "American Beauty."
} \\ \\
\parbox[c]{15.6cm}{
\textbf{MEMIT:} \underline{\textit{The name of the award the spouse of Melissa Mathison won is}} \textcolor{deepred}{not specified in the text}. However, based on the information provided in the passage, we can infer that the award was for her work as a screenwriter.
} \\ \\
\parbox[c]{15.6cm}{
\textbf{FT-L:} \underline{\textit{The name of the award the spouse of Melissa Mathison won is}} \textcolor{deepred}{I. A. Prize}. \textcolor{cyan}{nobody likes a snitch\textbackslash nI'm not sure what you are asking. Can you provide more context or clarify your question?}
} \\ \\
\parbox[c]{15.6cm}{
\textbf{LTE:} \underline{\textit{The name of the award the spouse of Melissa Mathison won is}} \textcolor{green}{I. I. Rabi Prize}. Melissa Mathison is married to Harrison Ford. The name of the award Harrison Ford won is \textcolor{green}{I. I. Rabi Prize}.
} \\ \bottomrule
\end{tabular}
\caption{Results for one case of different editing methods based on LLaMA2-Chat-7B. Queries are \underline{underlined} and \textit{italicized}. Words highlighted in \textcolor{green}{green} signify keywords that reflect correct behavior, while those in \textcolor{deepred}{red} denote keywords associated with incorrect behavior. Texts in \textcolor{cyan}{cyan} are repeated or meaningless sentences.}
\label{tab:case_study}
\end{table*}

\subsection{Case Study}

Table \ref{tab:case_study} shows the performance of different knowledge editing methods in a single case.
This comparison reveals that LTE stands out for applying edited knowledge to answer the query ``\texttt{The name of the award the spouse of Melissa Mathison won is}'' that necessitates compositional reasoning while maintaining the fluency of the generated text.
In contrast, other approaches, including SERAC, ICE, ROME, MEMIT, and FT-L, not only fail to answer the query correctly but generate repeated or inconsistent text sometimes.
This case study further validates LTE's advances in utilizing new knowledge to answer input queries.

\section{Related Work}

\paragraph{Knowledge Editing}
Prior knowledge editing studies resort to auxiliary models for efficient updating and refining of LLMs.
For example, SERAC~\cite{mitchell2022serac} builds a distinct counterfact model without changing the original LLM and employs a scope classifier to determine whether to use the counterfact model to answer the question.
KE~\cite{cao2021ke} and MEND~\cite{mitchell2022mend} leverage a hyper-network to predict the weight update of the LLM.
While these methods have shown some promising results, they fail to utilize the inherent formidable capabilities of LLMs.
More recent works such as KN~\cite{dai2022kn}, ROME~\cite{meng2022rome}, and MEMIT~\cite{meng2023memit} adopt interpretability techniques to identify parameters corresponding to specific knowledge and update them to alter LLM’s knowledge.
Nevertheless, the correlation between localization and editing efficacy has been questioned~\cite{hase2023does}.
Diverging from these methodologies, we explicitly teach LLMs how to apply updated knowledge rather than mere memorization, which taps into the full potential of LLMs, fostering a more dynamic and effective knowledge editing process.


\paragraph{LLM Alignment}
LLM alignment~\cite{gabriel2020artificial}, which aims to calibrate LLMs' behaviors with human values and preferences, is essential for their application in real-world scenarios.
A prominent technique in this area is supervised fine-tuning (SFT)~\cite{DBLP:conf/iclr/WeiBZGYLDDL22, DBLP:conf/acl/MishraKBH22}, which involves fine-tuning powerful LLMs using datasets composed of natural language instructions.
Notably, SFT is instrumental in improving LLMs’ understanding and adherence to human instructions, laying the groundwork for many subsequent alignment strategies such as reinforcement learning from human feedback (RLHF)~\cite{ouyang2022rlhf, rafailov2023direct}.
Thus, plenty of efforts have focused on applying SFT for alignment using either human-annotated or synthetic data~\cite{DBLP:conf/iclr/WeiBZGYLDDL22, wang2023selfinstruct, jiang2023lion, xu2023wizardlm, wang2023aligning}. 

\section{Conclusion}
We present the \textit{Learning to Edit} (LTE) framework, a novel approach for effective, efficient knowledge editing of LLMs. 
LTE equips LLMs with the ability to apply updated knowledge through a two-phase process: an Alignment Phase that teaches essential knowledge editing capabilities, and an Inference Phase that implements retrieval-based, on-the-fly knowledge editing.
Our framework demonstrates superior performance in knowledge editing tasks, outperforming existing methods in robustness and speed across various benchmarks.
\newpage

\section*{Limitations}
This section outlines the limitations of our proposed LTE approach, despite its validated efficacy across diverse model architectures, evaluation datasets, and knowledge editing settings.

Firstly, the LTE framework necessitates a one-time fine-tuning process during the Alignment Phase.
Although this process is a prerequisite, it facilitates real-time knowledge editing during the Inference Phase.
We further elucidate that employing LoRA as an alternative to standard fine-tuning presents a viable, resource-efficient approach without compromising performance (See \S \ref{sec:experiment}).
This innovation highlights the LTE's flexibility in adapting to various computational constraints.

Furthermore, our investigation primarily focuses on factual knowledge editing, yet the purview of model editing extends to encompassing personality traits, emotional responses, opinions, and beliefs~\cite{zhang2024knowedit}.
These dimensions, while partially explored, represent areas ripe for future research.
Additionally, the prospect of multilingual~\cite{DBLP:journals/corr/abs-2309-08952} and multimodal~\cite{cheng-etal-2023-edit} editing underscores the necessity for broader exploration, pointing towards an expansive horizon for model editing applications.

Finally, the proprietary nature of leading LLMs, such as ChatGPT and GPT-4, poses a significant challenge for applying knowledge editing techniques due to restricted access to their underlying parameters.
Nonetheless, OpenAI's API provision for models including \texttt{gpt-3.5-turbo-1106} and \texttt{gpt-4-0613} facilitates fine-tuning within the LTE's Alignment Phase. 
Although our current work does not extend to these black-box models, addressing this limitation represents a critical avenue for future research, potentially unlocking new methods for model customization and improvement.

\section*{Ethics Statement}

Knowledge editing encompasses the methodologies employed to efficiently modify LLMs’ behaviors within specific domains while preserving overall performance across various inputs.
It is critical to acknowledge that, if executed with malevolent intent, knowledge editing possesses the potential to compel models to produce outputs that are harmful or inappropriate.
Consequently, it is imperative to enforce safe and responsible knowledge editing practices.
The implementation of these techniques must be underpinned by stringent ethical standards, accompanied by robust safeguards designed to deter misuse and the generation of detrimental outcomes.
To this end, all data constructed for this study have undergone meticulous scrutiny by human evaluators to eliminate any instances of malicious editing or offensive content, thereby ensuring the integrity and safety of the model's outputs.

\section*{Acknowledgments}
W.\ Wang was supported by HKUST(GZ) Grant G0101000028, CCF-HuaweiDBC202302,  Guangzhou Municipal Science and Technology Project (No.\ 2023A03J0003, 2023A03J0013 and 2024A03J0621).

\newpage

\bibliography{custom}

\begin{thebibliography}{43}
\expandafter\ifx\csname natexlab\endcsname\relax\def\natexlab#1{#1}\fi

\bibitem[{Bai et~al.(2023)Bai, Bai, Chu, Cui, Dang, Deng, Fan, Ge, Han, Huang, Hui, Ji, Li, Lin, Lin, Liu, Liu, Lu, Lu, Ma, Men, Ren, Ren, Tan, Tan, Tu, Wang, Wang, Wang, Wu, Xu, Xu, Yang, Yang, Yang, Yang, Yao, Yu, Yuan, Yuan, Zhang, Zhang, Zhang, Zhang, Zhou, Zhou, Zhou, and Zhu}]{bai2023qwen}
Jinze Bai, Shuai Bai, Yunfei Chu, Zeyu Cui, Kai Dang, Xiaodong Deng, Yang Fan, Wenbin Ge, Yu~Han, Fei Huang, Binyuan Hui, Luo Ji, Mei Li, Junyang Lin, Runji Lin, Dayiheng Liu, Gao Liu, Chengqiang Lu, Keming Lu, Jianxin Ma, Rui Men, Xingzhang Ren, Xuancheng Ren, Chuanqi Tan, Sinan Tan, Jianhong Tu, Peng Wang, Shijie Wang, Wei Wang, Shengguang Wu, Benfeng Xu, Jin Xu, An~Yang, Hao Yang, Jian Yang, Shusheng Yang, Yang Yao, Bowen Yu, Hongyi Yuan, Zheng Yuan, Jianwei Zhang, Xingxuan Zhang, Yichang Zhang, Zhenru Zhang, Chang Zhou, Jingren Zhou, Xiaohuan Zhou, and Tianhang Zhu. 2023.
\newblock \href {https://doi.org/10.48550/ARXIV.2309.16609} {Qwen technical report}.
\newblock \emph{CoRR}, abs/2309.16609.

\bibitem[{Bisk et~al.(2020)Bisk, Zellers, Bras, Gao, and Choi}]{bisk2020piqa}
Yonatan Bisk, Rowan Zellers, Ronan~Le Bras, Jianfeng Gao, and Yejin Choi. 2020.
\newblock \href {https://doi.org/10.1609/AAAI.V34I05.6239} {{PIQA:} reasoning about physical commonsense in natural language}.
\newblock In \emph{The Thirty-Fourth {AAAI} Conference on Artificial Intelligence, {AAAI} 2020, The Thirty-Second Innovative Applications of Artificial Intelligence Conference, {IAAI} 2020, The Tenth {AAAI} Symposium on Educational Advances in Artificial Intelligence, {EAAI} 2020, New York, NY, USA, February 7-12, 2020}, pages 7432--7439. {AAAI} Press.

\bibitem[{Brown et~al.(2020)Brown, Mann, Ryder, Subbiah, Kaplan, Dhariwal, Neelakantan, Shyam, Sastry, Askell, Agarwal, Herbert{-}Voss, Krueger, Henighan, Child, Ramesh, Ziegler, Wu, Winter, Hesse, Chen, Sigler, Litwin, Gray, Chess, Clark, Berner, McCandlish, Radford, Sutskever, and Amodei}]{brown2020fewshot}
Tom~B. Brown, Benjamin Mann, Nick Ryder, Melanie Subbiah, Jared Kaplan, Prafulla Dhariwal, Arvind Neelakantan, Pranav Shyam, Girish Sastry, Amanda Askell, Sandhini Agarwal, Ariel Herbert{-}Voss, Gretchen Krueger, Tom Henighan, Rewon Child, Aditya Ramesh, Daniel~M. Ziegler, Jeffrey Wu, Clemens Winter, Christopher Hesse, Mark Chen, Eric Sigler, Mateusz Litwin, Scott Gray, Benjamin Chess, Jack Clark, Christopher Berner, Sam McCandlish, Alec Radford, Ilya Sutskever, and Dario Amodei. 2020.
\newblock \href {https://proceedings.neurips.cc/paper/2020/hash/1457c0d6bfcb4967418bfb8ac142f64a-Abstract.html} {Language models are few-shot learners}.
\newblock In \emph{Advances in Neural Information Processing Systems 33: Annual Conference on Neural Information Processing Systems 2020, NeurIPS 2020, December 6-12, 2020, virtual}.

\bibitem[{Chen et~al.(2021)Chen, Tworek, Jun, Yuan, de~Oliveira~Pinto, Kaplan, Edwards, Burda, Joseph, Brockman, Ray, Puri, Krueger, Petrov, Khlaaf, Sastry, Mishkin, Chan, Gray, Ryder, Pavlov, Power, Kaiser, Bavarian, Winter, Tillet, Such, Cummings, Plappert, Chantzis, Barnes, Herbert{-}Voss, Guss, Nichol, Paino, Tezak, Tang, Babuschkin, Balaji, Jain, Saunders, Hesse, Carr, Leike, Achiam, Misra, Morikawa, Radford, Knight, Brundage, Murati, Mayer, Welinder, McGrew, Amodei, McCandlish, Sutskever, and Zaremba}]{chen2021code}
Mark Chen, Jerry Tworek, Heewoo Jun, Qiming Yuan, Henrique~Pond{\'{e}} de~Oliveira~Pinto, Jared Kaplan, Harrison Edwards, Yuri Burda, Nicholas Joseph, Greg Brockman, Alex Ray, Raul Puri, Gretchen Krueger, Michael Petrov, Heidy Khlaaf, Girish Sastry, Pamela Mishkin, Brooke Chan, Scott Gray, Nick Ryder, Mikhail Pavlov, Alethea Power, Lukasz Kaiser, Mohammad Bavarian, Clemens Winter, Philippe Tillet, Felipe~Petroski Such, Dave Cummings, Matthias Plappert, Fotios Chantzis, Elizabeth Barnes, Ariel Herbert{-}Voss, William~Hebgen Guss, Alex Nichol, Alex Paino, Nikolas Tezak, Jie Tang, Igor Babuschkin, Suchir Balaji, Shantanu Jain, William Saunders, Christopher Hesse, Andrew~N. Carr, Jan Leike, Joshua Achiam, Vedant Misra, Evan Morikawa, Alec Radford, Matthew Knight, Miles Brundage, Mira Murati, Katie Mayer, Peter Welinder, Bob McGrew, Dario Amodei, Sam McCandlish, Ilya Sutskever, and Wojciech Zaremba. 2021.
\newblock \href {http://arxiv.org/abs/2107.03374} {Evaluating large language models trained on code}.
\newblock \emph{CoRR}, abs/2107.03374.

\bibitem[{Cheng et~al.(2023)Cheng, Tian, Liu, Chen, Wang, Chen, and Zhang}]{cheng-etal-2023-edit}
Siyuan Cheng, Bozhong Tian, Qingbin Liu, Xi~Chen, Yongheng Wang, Huajun Chen, and Ningyu Zhang. 2023.
\newblock \href {https://doi.org/10.18653/v1/2023.emnlp-main.856} {Can we edit multimodal large language models?}
\newblock In \emph{Proceedings of the 2023 Conference on Empirical Methods in Natural Language Processing}, pages 13877--13888, Singapore. Association for Computational Linguistics.

\bibitem[{Cohen et~al.(2023)Cohen, Biran, Yoran, Globerson, and Geva}]{cohen2024ice}
Roi Cohen, Eden Biran, Ori Yoran, Amir Globerson, and Mor Geva. 2023.
\newblock \href {https://doi.org/10.48550/ARXIV.2307.12976} {Evaluating the ripple effects of knowledge editing in language models}.
\newblock \emph{CoRR}, abs/2307.12976.

\bibitem[{Contributors(2023)}]{2023opencompass}
OpenCompass Contributors. 2023.
\newblock Opencompass: A universal evaluation platform for foundation models.
\newblock \url{https://github.com/open-compass/opencompass}.

\bibitem[{Dai et~al.(2022)Dai, Dong, Hao, Sui, Chang, and Wei}]{dai2022kn}
Damai Dai, Li~Dong, Yaru Hao, Zhifang Sui, Baobao Chang, and Furu Wei. 2022.
\newblock \href {https://doi.org/10.18653/V1/2022.ACL-LONG.581} {Knowledge neurons in pretrained transformers}.
\newblock In \emph{Proceedings of the 60th Annual Meeting of the Association for Computational Linguistics (Volume 1: Long Papers), {ACL} 2022, Dublin, Ireland, May 22-27, 2022}, pages 8493--8502. Association for Computational Linguistics.

\bibitem[{De~Cao et~al.(2021)De~Cao, Aziz, and Titov}]{cao2021ke}
Nicola De~Cao, Wilker Aziz, and Ivan Titov. 2021.
\newblock \href {https://doi.org/10.18653/v1/2021.emnlp-main.522} {Editing factual knowledge in language models}.
\newblock In \emph{Proceedings of the 2021 Conference on Empirical Methods in Natural Language Processing}, pages 6491--6506, Online and Punta Cana, Dominican Republic. Association for Computational Linguistics.

\bibitem[{Gabriel(2020)}]{gabriel2020artificial}
Iason Gabriel. 2020.
\newblock \href {https://doi.org/10.1007/S11023-020-09539-2} {Artificial intelligence, values, and alignment}.
\newblock \emph{Minds Mach.}, 30(3):411--437.

\bibitem[{Hartvigsen et~al.(2023)Hartvigsen, Sankaranarayanan, Palangi, Kim, and Ghassemi}]{hartvigsen2023grace}
Thomas Hartvigsen, Swami Sankaranarayanan, Hamid Palangi, Yoon Kim, and Marzyeh Ghassemi. 2023.
\newblock Aging with grace: Lifelong model editing with discrete key-value adaptors.
\newblock In \emph{Advances in Neural Information Processing Systems}.

\bibitem[{Hase et~al.(2023)Hase, Bansal, Kim, and Ghandeharioun}]{hase2023does}
Peter Hase, Mohit Bansal, Been Kim, and Asma Ghandeharioun. 2023.
\newblock \href {http://arxiv.org/abs/2301.04213} {Does localization inform editing? surprising differences in causality-based localization vs. knowledge editing in language models}.

\bibitem[{Hendrycks et~al.(2021)Hendrycks, Burns, Basart, Zou, Mazeika, Song, and Steinhardt}]{dan2021mmlu}
Dan Hendrycks, Collin Burns, Steven Basart, Andy Zou, Mantas Mazeika, Dawn Song, and Jacob Steinhardt. 2021.
\newblock \href {https://openreview.net/forum?id=d7KBjmI3GmQ} {Measuring massive multitask language understanding}.
\newblock In \emph{9th International Conference on Learning Representations, {ICLR} 2021, Virtual Event, Austria, May 3-7, 2021}. OpenReview.net.

\bibitem[{Hu et~al.(2022)Hu, Shen, Wallis, Allen-Zhu, Li, Wang, Wang, and Chen}]{hu2022lora}
Edward~J Hu, Yelong Shen, Phillip Wallis, Zeyuan Allen-Zhu, Yuanzhi Li, Shean Wang, Lu~Wang, and Weizhu Chen. 2022.
\newblock \href {https://openreview.net/forum?id=nZeVKeeFYf9} {Lo{RA}: Low-rank adaptation of large language models}.
\newblock In \emph{International Conference on Learning Representations}.

\bibitem[{Jiang et~al.(2023)Jiang, Chan, Chen, and Wang}]{jiang2023lion}
Yuxin Jiang, Chunkit Chan, Mingyang Chen, and Wei Wang. 2023.
\newblock \href {https://doi.org/10.18653/v1/2023.emnlp-main.189} {Lion: Adversarial distillation of proprietary large language models}.
\newblock In \emph{Proceedings of the 2023 Conference on Empirical Methods in Natural Language Processing}, pages 3134--3154, Singapore. Association for Computational Linguistics.

\bibitem[{Levy et~al.(2017)Levy, Seo, Choi, and Zettlemoyer}]{levy2017zsre}
Omer Levy, Minjoon Seo, Eunsol Choi, and Luke Zettlemoyer. 2017.
\newblock \href {https://doi.org/10.18653/V1/K17-1034} {Zero-shot relation extraction via reading comprehension}.
\newblock In \emph{Proceedings of the 21st Conference on Computational Natural Language Learning (CoNLL 2017), Vancouver, Canada, August 3-4, 2017}, pages 333--342. Association for Computational Linguistics.

\bibitem[{Lewis et~al.(2020)Lewis, Perez, Piktus, Petroni, Karpukhin, Goyal, K{\"{u}}ttler, Lewis, Yih, Rockt{\"{a}}schel, Riedel, and Kiela}]{lewis2020rag}
Patrick S.~H. Lewis, Ethan Perez, Aleksandra Piktus, Fabio Petroni, Vladimir Karpukhin, Naman Goyal, Heinrich K{\"{u}}ttler, Mike Lewis, Wen{-}tau Yih, Tim Rockt{\"{a}}schel, Sebastian Riedel, and Douwe Kiela. 2020.
\newblock \href {https://proceedings.neurips.cc/paper/2020/hash/6b493230205f780e1bc26945df7481e5-Abstract.html} {Retrieval-augmented generation for knowledge-intensive {NLP} tasks}.
\newblock In \emph{Advances in Neural Information Processing Systems 33: Annual Conference on Neural Information Processing Systems 2020, NeurIPS 2020, December 6-12, 2020, virtual}.

\bibitem[{Li et~al.(2023)Li, Zhang, Dubois, Taori, Gulrajani, Guestrin, Liang, and Hashimoto}]{alpaca_eval}
Xuechen Li, Tianyi Zhang, Yann Dubois, Rohan Taori, Ishaan Gulrajani, Carlos Guestrin, Percy Liang, and Tatsunori~B. Hashimoto. 2023.
\newblock Alpacaeval: An automatic evaluator of instruction-following models.
\newblock \url{https://github.com/tatsu-lab/alpaca_eval}.

\bibitem[{Meng et~al.(2022)Meng, Bau, Andonian, and Belinkov}]{meng2022rome}
Kevin Meng, David Bau, Alex Andonian, and Yonatan Belinkov. 2022.
\newblock \href {http://papers.nips.cc/paper\_files/paper/2022/hash/6f1d43d5a82a37e89b0665b33bf3a182-Abstract-Conference.html} {Locating and editing factual associations in {GPT}}.
\newblock In \emph{Advances in Neural Information Processing Systems 35: Annual Conference on Neural Information Processing Systems 2022, NeurIPS 2022, New Orleans, LA, USA, November 28 - December 9, 2022}.

\bibitem[{Meng et~al.(2023)Meng, Sharma, Andonian, Belinkov, and Bau}]{meng2023memit}
Kevin Meng, Arnab~Sen Sharma, Alex~J. Andonian, Yonatan Belinkov, and David Bau. 2023.
\newblock \href {https://openreview.net/pdf?id=MkbcAHIYgyS} {Mass-editing memory in a transformer}.
\newblock In \emph{The Eleventh International Conference on Learning Representations, {ICLR} 2023, Kigali, Rwanda, May 1-5, 2023}. OpenReview.net.

\bibitem[{Mishra et~al.(2022)Mishra, Khashabi, Baral, and Hajishirzi}]{DBLP:conf/acl/MishraKBH22}
Swaroop Mishra, Daniel Khashabi, Chitta Baral, and Hannaneh Hajishirzi. 2022.
\newblock \href {https://doi.org/10.18653/v1/2022.acl-long.244} {Cross-task generalization via natural language crowdsourcing instructions}.
\newblock In \emph{Proceedings of the 60th Annual Meeting of the Association for Computational Linguistics (Volume 1: Long Papers), {ACL} 2022, Dublin, Ireland, May 22-27, 2022}, pages 3470--3487. Association for Computational Linguistics.

\bibitem[{Mitchell et~al.(2022{\natexlab{a}})Mitchell, Lin, Bosselut, Finn, and Manning}]{mitchell2022mend}
Eric Mitchell, Charles Lin, Antoine Bosselut, Chelsea Finn, and Christopher~D. Manning. 2022{\natexlab{a}}.
\newblock \href {https://openreview.net/forum?id=0DcZxeWfOPt} {Fast model editing at scale}.
\newblock In \emph{The Tenth International Conference on Learning Representations, {ICLR} 2022, Virtual Event, April 25-29, 2022}. OpenReview.net.

\bibitem[{Mitchell et~al.(2022{\natexlab{b}})Mitchell, Lin, Bosselut, Manning, and Finn}]{mitchell2022serac}
Eric Mitchell, Charles Lin, Antoine Bosselut, Christopher~D. Manning, and Chelsea Finn. 2022{\natexlab{b}}.
\newblock \href {https://proceedings.mlr.press/v162/mitchell22a.html} {Memory-based model editing at scale}.
\newblock In \emph{International Conference on Machine Learning, {ICML} 2022, 17-23 July 2022, Baltimore, Maryland, {USA}}, volume 162 of \emph{Proceedings of Machine Learning Research}, pages 15817--15831. {PMLR}.

\bibitem[{Narayan et~al.(2018)Narayan, Cohen, and Lapata}]{narayan2018xsum}
Shashi Narayan, Shay~B. Cohen, and Mirella Lapata. 2018.
\newblock \href {https://doi.org/10.18653/V1/D18-1206} {Don't give me the details, just the summary! topic-aware convolutional neural networks for extreme summarization}.
\newblock In \emph{Proceedings of the 2018 Conference on Empirical Methods in Natural Language Processing, Brussels, Belgium, October 31 - November 4, 2018}, pages 1797--1807. Association for Computational Linguistics.

\bibitem[{OpenAI(2023)}]{2023gpt4}
OpenAI. 2023.
\newblock \href {http://arxiv.org/abs/2303.08774} {{GPT-4} technical report}.
\newblock \emph{CoRR}, abs/2303.08774.

\bibitem[{OpenAI(2022)}]{openai2022chatgpt}
TB~OpenAI. 2022.
\newblock Chatgpt: Optimizing language models for dialogue.
\newblock \emph{OpenAI}.

\bibitem[{Ouyang et~al.(2022)Ouyang, Wu, Jiang, Almeida, Wainwright, Mishkin, Zhang, Agarwal, Slama, Ray, Schulman, Hilton, Kelton, Miller, Simens, Askell, Welinder, Christiano, Leike, and Lowe}]{ouyang2022rlhf}
Long Ouyang, Jeffrey Wu, Xu~Jiang, Diogo Almeida, Carroll~L. Wainwright, Pamela Mishkin, Chong Zhang, Sandhini Agarwal, Katarina Slama, Alex Ray, John Schulman, Jacob Hilton, Fraser Kelton, Luke Miller, Maddie Simens, Amanda Askell, Peter Welinder, Paul~F. Christiano, Jan Leike, and Ryan Lowe. 2022.
\newblock \href {http://papers.nips.cc/paper\_files/paper/2022/hash/b1efde53be364a73914f58805a001731-Abstract-Conference.html} {Training language models to follow instructions with human feedback}.
\newblock In \emph{Advances in Neural Information Processing Systems 35: Annual Conference on Neural Information Processing Systems 2022, NeurIPS 2022, New Orleans, LA, USA, November 28 - December 9, 2022}.

\bibitem[{Rafailov et~al.(2023)Rafailov, Sharma, Mitchell, Manning, Ermon, and Finn}]{rafailov2023direct}
Rafael Rafailov, Archit Sharma, Eric Mitchell, Christopher~D Manning, Stefano Ermon, and Chelsea Finn. 2023.
\newblock \href {https://arxiv.org/abs/2305.18290} {Direct preference optimization: Your language model is secretly a reward model}.
\newblock In \emph{Thirty-seventh Conference on Neural Information Processing Systems}.

\bibitem[{Reimers and Gurevych(2019)}]{reimers2019sentencebert}
Nils Reimers and Iryna Gurevych. 2019.
\newblock \href {https://arxiv.org/abs/1908.10084} {Sentence-bert: Sentence embeddings using siamese bert-networks}.
\newblock In \emph{Proceedings of the 2019 Conference on Empirical Methods in Natural Language Processing}. Association for Computational Linguistics.

\bibitem[{Sinitsin et~al.(2020)Sinitsin, Plokhotnyuk, Pyrkin, Popov, and Babenko}]{DBLP:conf/iclr/SinitsinPPPB20}
Anton Sinitsin, Vsevolod Plokhotnyuk, Dmitry~V. Pyrkin, Sergei Popov, and Artem Babenko. 2020.
\newblock \href {https://openreview.net/forum?id=HJedXaEtvS} {Editable neural networks}.
\newblock In \emph{8th International Conference on Learning Representations, {ICLR} 2020, Addis Ababa, Ethiopia, April 26-30, 2020}. OpenReview.net.

\bibitem[{Talmor et~al.(2019)Talmor, Herzig, Lourie, and Berant}]{talmor2019commonsenseqa}
Alon Talmor, Jonathan Herzig, Nicholas Lourie, and Jonathan Berant. 2019.
\newblock \href {https://doi.org/10.18653/v1/N19-1421} {{C}ommonsense{QA}: A question answering challenge targeting commonsense knowledge}.
\newblock In \emph{Proceedings of the 2019 Conference of the North {A}merican Chapter of the Association for Computational Linguistics: Human Language Technologies, Volume 1 (Long and Short Papers)}, pages 4149--4158, Minneapolis, Minnesota. Association for Computational Linguistics.

\bibitem[{Touvron et~al.(2023)Touvron, Martin, Stone, Albert, Almahairi, Babaei, Bashlykov, Batra, Bhargava, Bhosale, Bikel, Blecher, Ferrer, Chen, Cucurull, Esiobu, Fernandes, Fu, Fu, Fuller, Gao, Goswami, Goyal, Hartshorn, Hosseini, Hou, Inan, Kardas, Kerkez, Khabsa, Kloumann, Korenev, Koura, Lachaux, Lavril, Lee, Liskovich, Lu, Mao, Martinet, Mihaylov, Mishra, Molybog, Nie, Poulton, Reizenstein, Rungta, Saladi, Schelten, Silva, Smith, Subramanian, Tan, Tang, Taylor, Williams, Kuan, Xu, Yan, Zarov, Zhang, Fan, Kambadur, Narang, Rodriguez, Stojnic, Edunov, and Scialom}]{touvron2023llama}
Hugo Touvron, Louis Martin, Kevin Stone, Peter Albert, Amjad Almahairi, Yasmine Babaei, Nikolay Bashlykov, Soumya Batra, Prajjwal Bhargava, Shruti Bhosale, Dan Bikel, Lukas Blecher, Cristian~Canton Ferrer, Moya Chen, Guillem Cucurull, David Esiobu, Jude Fernandes, Jeremy Fu, Wenyin Fu, Brian Fuller, Cynthia Gao, Vedanuj Goswami, Naman Goyal, Anthony Hartshorn, Saghar Hosseini, Rui Hou, Hakan Inan, Marcin Kardas, Viktor Kerkez, Madian Khabsa, Isabel Kloumann, Artem Korenev, Punit~Singh Koura, Marie-Anne Lachaux, Thibaut Lavril, Jenya Lee, Diana Liskovich, Yinghai Lu, Yuning Mao, Xavier Martinet, Todor Mihaylov, Pushkar Mishra, Igor Molybog, Yixin Nie, Andrew Poulton, Jeremy Reizenstein, Rashi Rungta, Kalyan Saladi, Alan Schelten, Ruan Silva, Eric~Michael Smith, Ranjan Subramanian, Xiaoqing~Ellen Tan, Binh Tang, Ross Taylor, Adina Williams, Jian~Xiang Kuan, Puxin Xu, Zheng Yan, Iliyan Zarov, Yuchen Zhang, Angela Fan, Melanie Kambadur, Sharan Narang, Aurelien Rodriguez, Robert Stojnic, Sergey Edunov, and Thomas
  Scialom. 2023.
\newblock \href {http://arxiv.org/abs/2307.09288} {Llama 2: Open foundation and fine-tuned chat models}.

\bibitem[{Wang et~al.(2023{\natexlab{a}})Wang, Liang, Sun, Cao, and Xu}]{DBLP:journals/corr/abs-2309-08952}
Jiaan Wang, Yunlong Liang, Zengkui Sun, Yuxuan Cao, and Jiarong Xu. 2023{\natexlab{a}}.
\newblock \href {https://doi.org/10.48550/ARXIV.2309.08952} {Cross-lingual knowledge editing in large language models}.
\newblock \emph{CoRR}, abs/2309.08952.

\bibitem[{Wang et~al.(2023{\natexlab{b}})Wang, Zhang, Xie, Yao, Tian, Wang, Xi, Cheng, Liu, Zheng, and Chen}]{wang2023easyedit}
Peng Wang, Ningyu Zhang, Xin Xie, Yunzhi Yao, Bozhong Tian, Mengru Wang, Zekun Xi, Siyuan Cheng, Kangwei Liu, Guozhou Zheng, and Huajun Chen. 2023{\natexlab{b}}.
\newblock \href {https://doi.org/10.48550/ARXIV.2308.07269} {Easyedit: An easy-to-use knowledge editing framework for large language models}.
\newblock \emph{CoRR}, abs/2308.07269.

\bibitem[{Wang et~al.(2023{\natexlab{c}})Wang, Kordi, Mishra, Liu, Smith, Khashabi, and Hajishirzi}]{wang2023selfinstruct}
Yizhong Wang, Yeganeh Kordi, Swaroop Mishra, Alisa Liu, Noah~A. Smith, Daniel Khashabi, and Hannaneh Hajishirzi. 2023{\natexlab{c}}.
\newblock \href {https://doi.org/10.18653/v1/2023.acl-long.754} {Self-instruct: Aligning language models with self-generated instructions}.
\newblock In \emph{Proceedings of the 61st Annual Meeting of the Association for Computational Linguistics (Volume 1: Long Papers)}, pages 13484--13508, Toronto, Canada. Association for Computational Linguistics.

\bibitem[{Wang et~al.(2023{\natexlab{d}})Wang, Zhong, Li, Mi, Zeng, Huang, Shang, Jiang, and Liu}]{wang2023aligning}
Yufei Wang, Wanjun Zhong, Liangyou Li, Fei Mi, Xingshan Zeng, Wenyong Huang, Lifeng Shang, Xin Jiang, and Qun Liu. 2023{\natexlab{d}}.
\newblock Aligning large language models with human: A survey.
\newblock \emph{arXiv preprint arXiv:2307.12966}.

\bibitem[{Wei et~al.(2022)Wei, Bosma, Zhao, Guu, Yu, Lester, Du, Dai, and Le}]{DBLP:conf/iclr/WeiBZGYLDDL22}
Jason Wei, Maarten Bosma, Vincent~Y. Zhao, Kelvin Guu, Adams~Wei Yu, Brian Lester, Nan Du, Andrew~M. Dai, and Quoc~V. Le. 2022.
\newblock \href {https://openreview.net/forum?id=gEZrGCozdqR} {Finetuned language models are zero-shot learners}.
\newblock In \emph{The Tenth International Conference on Learning Representations, {ICLR} 2022, Virtual Event, April 25-29, 2022}. OpenReview.net.

\bibitem[{Xu et~al.(2023)Xu, Sun, Zheng, Geng, Zhao, Feng, Tao, and Jiang}]{xu2023wizardlm}
Can Xu, Qingfeng Sun, Kai Zheng, Xiubo Geng, Pu~Zhao, Jiazhan Feng, Chongyang Tao, and Daxin Jiang. 2023.
\newblock \href {https://doi.org/10.48550/ARXIV.2304.12244} {Wizardlm: Empowering large language models to follow complex instructions}.
\newblock \emph{CoRR}, abs/2304.12244.

\bibitem[{Xu et~al.(2022)Xu, Szlam, and Weston}]{xu2022rag}
Jing Xu, Arthur Szlam, and Jason Weston. 2022.
\newblock \href {https://doi.org/10.18653/V1/2022.ACL-LONG.356} {Beyond goldfish memory: Long-term open-domain conversation}.
\newblock In \emph{Proceedings of the 60th Annual Meeting of the Association for Computational Linguistics (Volume 1: Long Papers), {ACL} 2022, Dublin, Ireland, May 22-27, 2022}, pages 5180--5197. Association for Computational Linguistics.

\bibitem[{Zhang et~al.(2024)Zhang, Yao, Tian, Wang, Deng, Wang, Xi, Mao, Zhang, Ni, Cheng, Xu, Xu, Gu, Jiang, Xie, Huang, Liang, Zhang, Zhu, Zhou, and Chen}]{zhang2024knowedit}
Ningyu Zhang, Yunzhi Yao, Bozhong Tian, Peng Wang, Shumin Deng, Mengru Wang, Zekun Xi, Shengyu Mao, Jintian Zhang, Yuansheng Ni, Siyuan Cheng, Ziwen Xu, Xin Xu, Jia{-}Chen Gu, Yong Jiang, Pengjun Xie, Fei Huang, Lei Liang, Zhiqiang Zhang, Xiaowei Zhu, Jun Zhou, and Huajun Chen. 2024.
\newblock \href {https://doi.org/10.48550/ARXIV.2401.01286} {A comprehensive study of knowledge editing for large language models}.
\newblock \emph{CoRR}, abs/2401.01286.

\bibitem[{Zhang et~al.(2018)Zhang, Galley, Gao, Gan, Li, Brockett, and Dolan}]{zhang2018generating}
Yizhe Zhang, Michel Galley, Jianfeng Gao, Zhe Gan, Xiujun Li, Chris Brockett, and Bill Dolan. 2018.
\newblock Generating informative and diverse conversational responses via adversarial information maximization.
\newblock \emph{Advances in Neural Information Processing Systems}, 31.

\bibitem[{Zhong et~al.(2023{\natexlab{a}})Zhong, Cui, Guo, Liang, Lu, Wang, Saied, Chen, and Duan}]{zhong2023agieval}
Wanjun Zhong, Ruixiang Cui, Yiduo Guo, Yaobo Liang, Shuai Lu, Yanlin Wang, Amin Saied, Weizhu Chen, and Nan Duan. 2023{\natexlab{a}}.
\newblock \href {https://doi.org/10.48550/ARXIV.2304.06364} {Agieval: {A} human-centric benchmark for evaluating foundation models}.
\newblock \emph{CoRR}, abs/2304.06364.

\bibitem[{Zhong et~al.(2023{\natexlab{b}})Zhong, Wu, Manning, Potts, and Chen}]{zhong2023mquake}
Zexuan Zhong, Zhengxuan Wu, Christopher~D. Manning, Christopher Potts, and Danqi Chen. 2023{\natexlab{b}}.
\newblock \href {https://aclanthology.org/2023.emnlp-main.971} {Mquake: Assessing knowledge editing in language models via multi-hop questions}.
\newblock In \emph{Proceedings of the 2023 Conference on Empirical Methods in Natural Language Processing, {EMNLP} 2023, Singapore, December 6-10, 2023}, pages 15686--15702. Association for Computational Linguistics.

\end{thebibliography}
\bibliographystyle{acl_natbib}

\appendix

\section{Details of Training Data Construction}
\label{sec:appendix_data_construction}

\subsection{Synthetics of Out-of-scope Examples}
\label{sec:appendix_out_scope}
\begin{figure*}[!t]
\centering
\includegraphics[width=\linewidth]{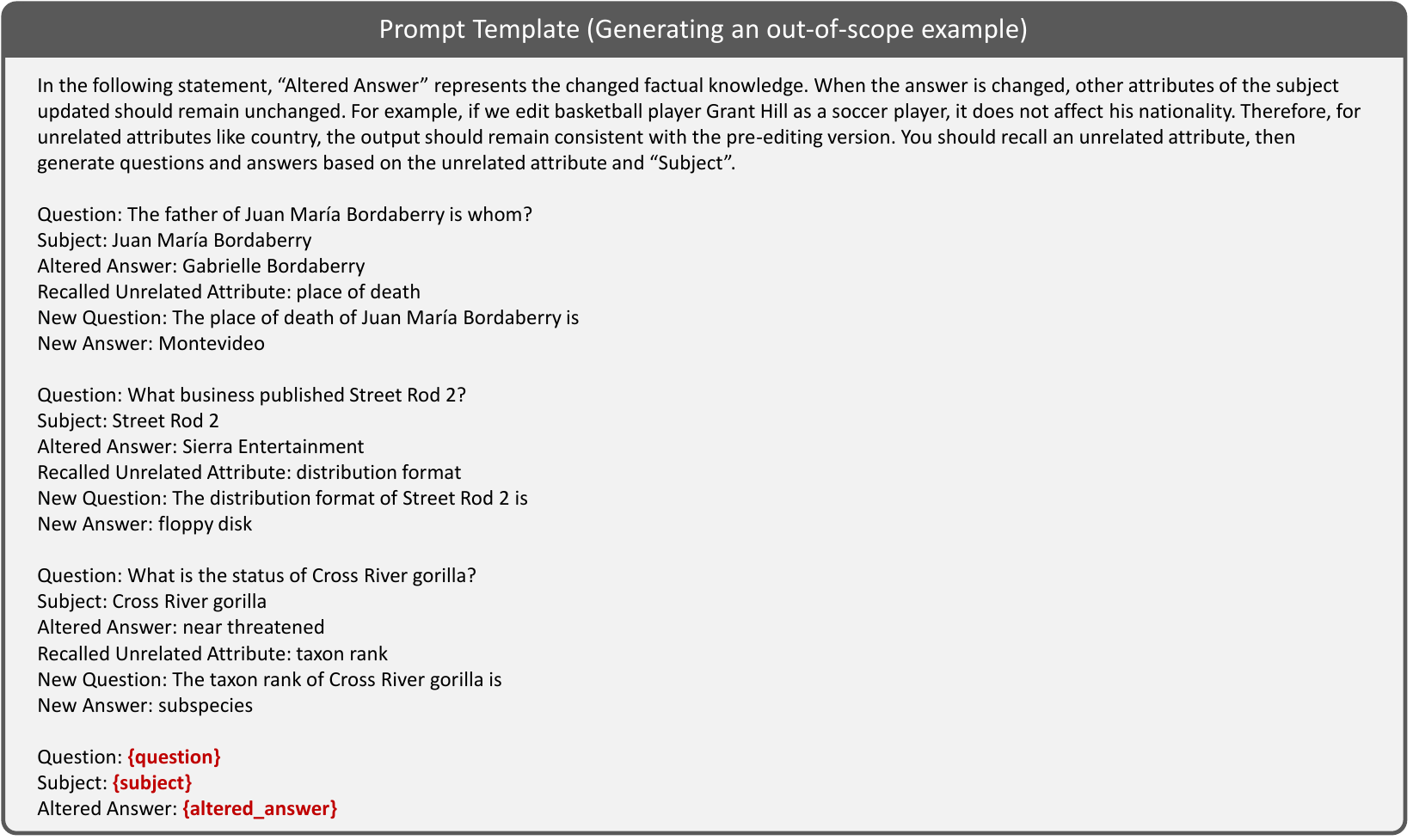}
\caption{
Prompt template for generating an out-of-scope example.
}
\label{fig:example_prompt}
\end{figure*}

As shown in Figure \ref{fig:example_prompt}, we employ a few-shot manual demonstration as a prompt to guide GPT-4 in producing the desired query and answer.

\begin{table*}[!t]
\footnotesize
\centering
\begin{tabularx}{\textwidth}{lCCCCcc}
\toprule
\textbf{Data Source} & \parbox{1.8cm}{\textbf{\# of in-scope; \\ w/ prompt}} & \parbox{1.8cm}{\textbf{\# of in-scope;\\ w/o prompt}} & \parbox{2.3cm}{\textbf{\# of out-of-scope;\\ w/ prompt}} & \parbox{2.3cm}{\textbf{\# of out-of-scope;\\ w/o prompt}} & \textbf{\# of Total} & \textbf{Avg Len} \\ \midrule
ZsRE        & 1,000                     & 1,000                      & 1,000                         & 1,000                          & 4,000       &  27         \\
RIPPLEEDITS & 2,250                     & 2,250                      & 2,250                         & 2,250                          & 9,000       & 34          \\
WikiBio     & 250                       & 250                        & 250                           & 250                            & 1,000       & 102          \\
MQUAKE      & 4,000                     & 4,000                      & 4,000                         & 4,000                          & 16,000      &  160         \\
COUNTERFACT & 7,500                     & 7,500                      & 7,500                         & 7,500                          & 30,000      & 320          \\ \midrule
Total       & 15,000                    & 15,000                     & 15,000                        & 15,000                         & 60,000      & 208         \\ \bottomrule
\end{tabularx}
\caption{Training data statistics. ``Avg Len'' is the average word number of samples, and ``prompt'' denotes our designed knowledge editing prompt template in Figure \ref{fig:method}.}
\label{tab:statistics}
\end{table*}

\subsection{Synthetics of Free-text In-scope Question-answering Pairs}
\label{sec:appendix_free_text}
\begin{figure*}[!t]
\centering
\includegraphics[width=\linewidth]{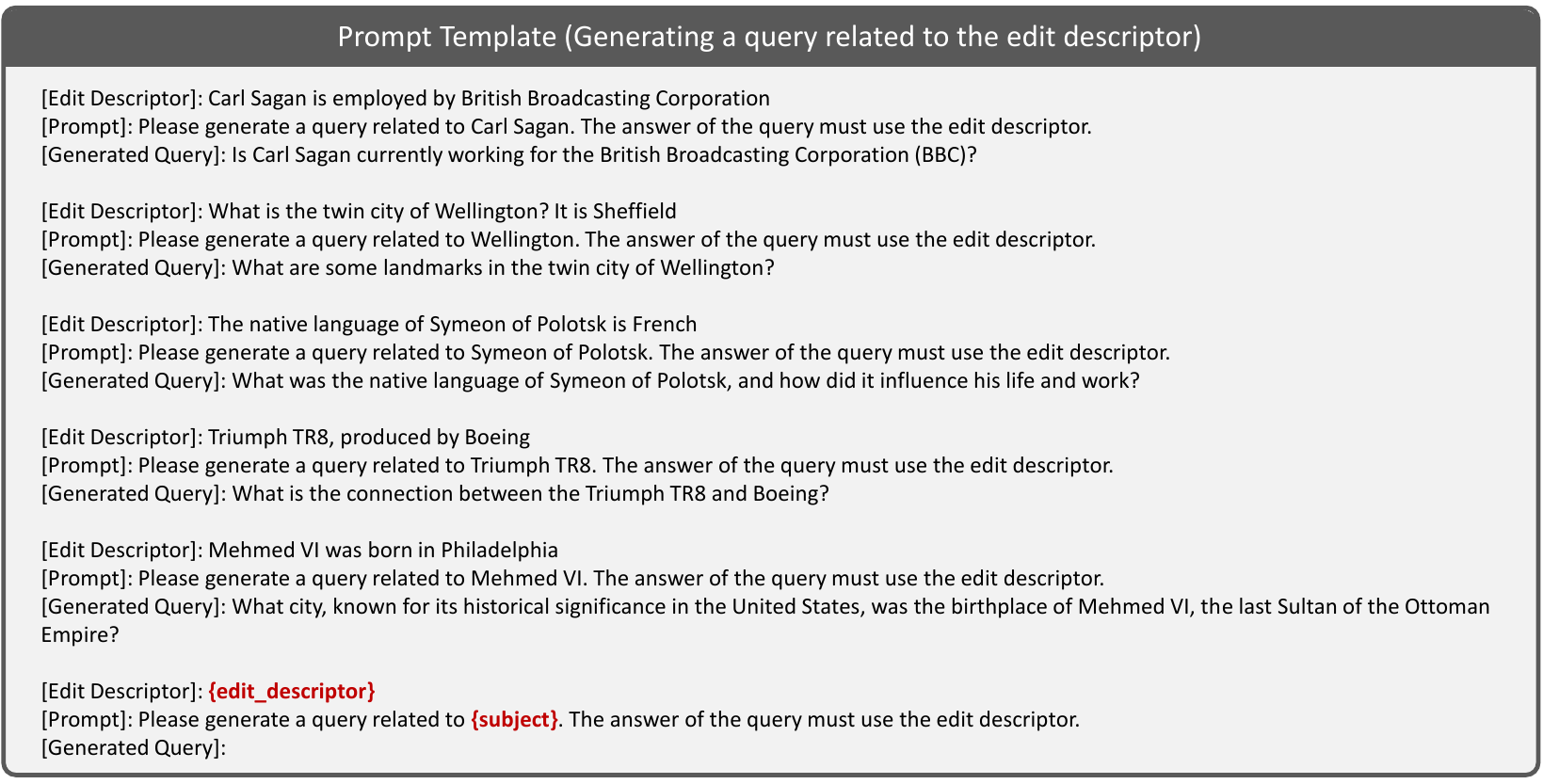}
\caption{
Prompt template for generating a query related to the edit descriptor.
}
\label{fig:question_prompt}
\end{figure*}

\begin{figure*}[!t]
\centering
\includegraphics[width=\linewidth]{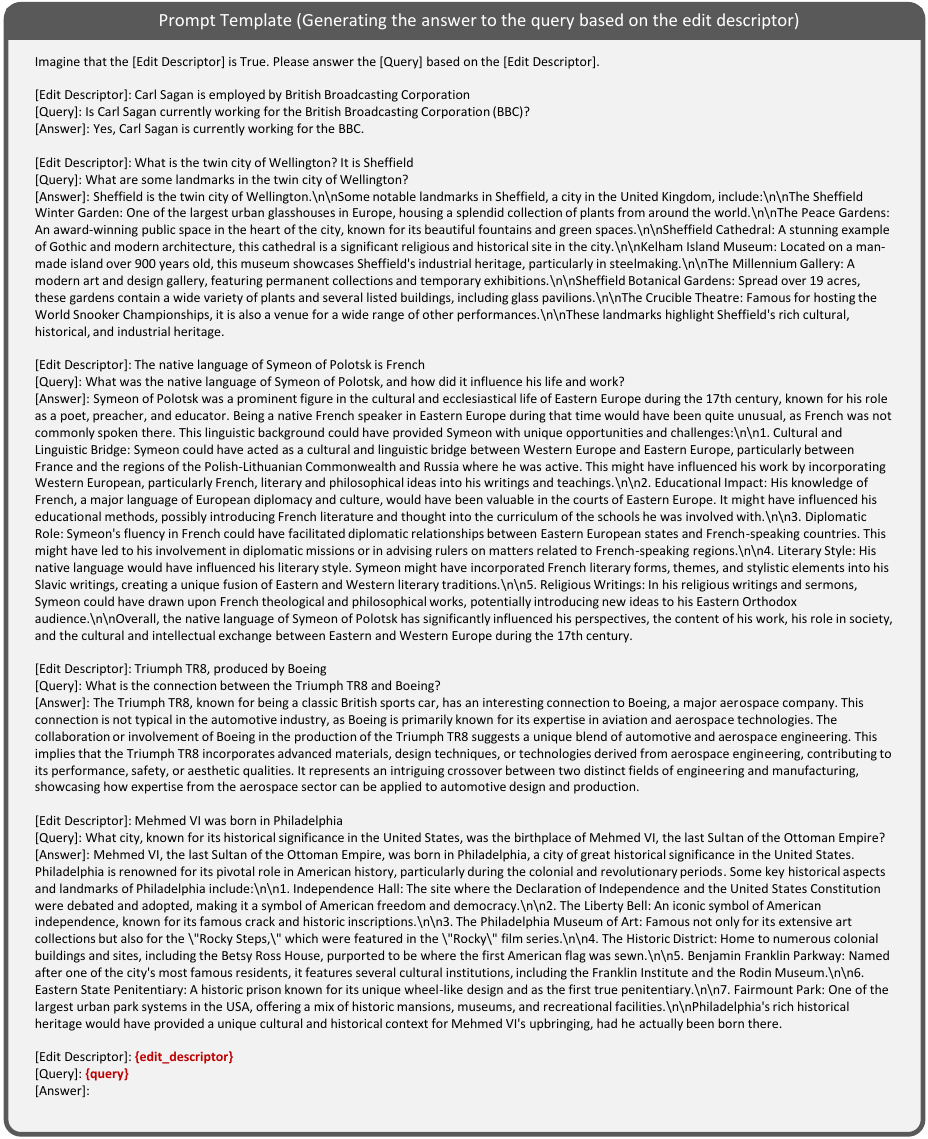}
\caption{
Prompt template for generating the answer to the query based on the edit descriptor.
}
\label{fig:answer_prompt}
\end{figure*}

\begin{figure*}[!t]
\centering
\includegraphics[width=\linewidth]{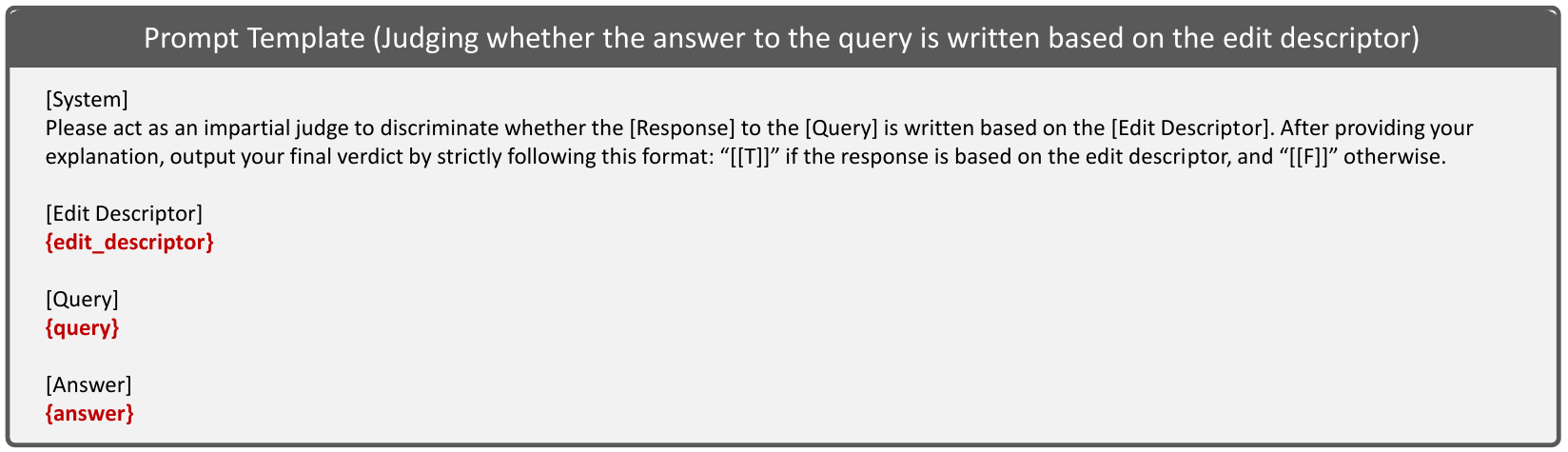}
\caption{
Prompt template for judging whether the answer to the query is written based on the edit descriptor.
}
\label{fig:check_prompt}
\end{figure*}

In our methodology, we initially engage GPT-4 with five meticulously crafted demonstrations, as depicted in Figure \ref{fig:question_prompt}.
This step is designed to elicit a query that pertains directly to the edit descriptor.
Following this, we direct GPT-4 to formulate an answer to the query, drawing upon the edit descriptor for content, as illustrated in Figure \ref{fig:answer_prompt}. 
The final step in Figure \ref{fig:check_prompt} involves a verification process by GPT-4 to ascertain the congruence of the answer with the edit descriptor, leading to the exclusion of instances where the criteria are not met (approximately 15\%).

\subsection{Training Data Statistics}
\label{sec:appendix_statistics}
Table \ref{tab:statistics} lists the statistics of our curated training data, which encompasses 60k samples from five data sources.
In the construction of our dataset, we employ a rigorous sampling methodology, exclusively selecting instances from the training sets provided by the data sources.


\section{Implementation Details}
\label{sec:appendix_implementation}
The training procedure was executed on 4 NVIDIA A100 GPUs, each equipped with 80GB of memory. The duration required to train a single instance of the model, specifically the LLaMA2-Chat-7B, was approximately 9 hours. Detailed specifications of the hyperparameters employed for both standard fine-tuning and LoRA are provided in Table \ref{tab:hyperparameters}.

\begin{table}[!h]
\small
\centering
{\begin{tabularx}{\linewidth}{l|C|C}
\toprule
\textbf{Hyperparameter} & \textbf{Standard FT} & \textbf{LoRA} \\ \midrule
Batch size              & 128              & 128               \\
Learning rate           & 2e-5             & 3e-4              \\
Epoches                 & 3                & 3                 \\
Max length              & 2048             & 2048              \\
Optimizer               & AdamW            & AdamW             \\
Scheduler               & cosine           & cosine            \\
Weight decay            & 0                & 0                 \\
Warmup ratio            & 0.03             & 0.03           \\ \bottomrule
\end{tabularx}}
\caption{\label{tab:hyperparameters}
Training hyperparameters for both LLaMA2-Chat-7B and Qwen-Chat-7B.}
\end{table}

\end{document}